\documentclass[10pt,twocolumn,letterpaper]{article}

\usepackage{wacv}
\usepackage{times}
\usepackage{epsfig}
\usepackage{graphicx}
\usepackage{amsmath}
\usepackage{amssymb}
\usepackage{booktabs}
\usepackage{siunitx}
\usepackage{booktabs}
\usepackage{multirow}
\usepackage{makecell}
\usepackage{xspace}
\usepackage{multicol}
\usepackage{vwcol}
\usepackage{microtype}
\usepackage[accsupp]{axessibility}  %

\usepackage{tabularx}
\newcolumntype{Y}{>{\centering\arraybackslash}X}

\def\wacvPaperID{430} %

\wacvfinalcopy %

\ifwacvfinal
\usepackage[breaklinks=true,bookmarks=false]{hyperref}
\else
\usepackage[pagebackref=true,breaklinks=true,colorlinks,bookmarks=false]{hyperref}
\fi

\usepackage{cleveref}

\newcommand{\model}{InDiReCT\xspace}
\newcommand{\longmodel}{\textbf{I}mage representatio\textbf{n}s using \textbf{Di}men\-sionality \textbf{Re}duction on \textbf{C}LIP embedded \textbf{T}exts\xspace}
\newcommand{\task}{LanZ-DML\xspace}
\newcommand{\longtask}{Language-Guided Zero-Shot Deep Metric Learning\xspace}

\begin{document}

\title{\model: \longtask for Images}

\author{Konstantin Kobs \qquad Michael Steininger \qquad Andreas Hotho \\
University of Würzburg\\
Am Hubland, 97074 Würzburg\\
{\tt\small \{kobs,steininger,hotho\}@informatik.uni-wuerzburg.de}
}

\maketitle
\thispagestyle{empty}

\begin{abstract}
    Common Deep Metric Learning (DML) datasets specify only one notion of similarity, e.g., two images in the Cars196 dataset are deemed similar if they show the same car model.
    We argue that depending on the application, users of image retrieval systems have different and changing similarity notions that should be incorporated as easily as possible.
    Therefore, we present \textit{\longtask (\task)} as a new DML setting in which users control the properties that should be important for image representations without training data by only using natural language.
    To this end, we propose \textit{\model (\longmodel)}, a model for \task on images that exclusively uses a few text prompts for training.
    \model utilizes CLIP as a fixed feature extractor for images and texts and transfers the variation in text prompt embeddings to the image embedding space.
    Extensive experiments on five datasets and overall thirteen similarity notions show that, despite not seeing any images during training, \model performs better than strong baselines and approaches the performance of fully-supervised models.
    An analysis reveals that \model learns to focus on regions of the image that correlate with the desired similarity notion, which makes it a fast to train and easy to use method to create custom embedding spaces only using natural language.
\end{abstract}

\section{Introduction}

\begin{figure}[t]
    \centering
    \includegraphics[width=\linewidth]{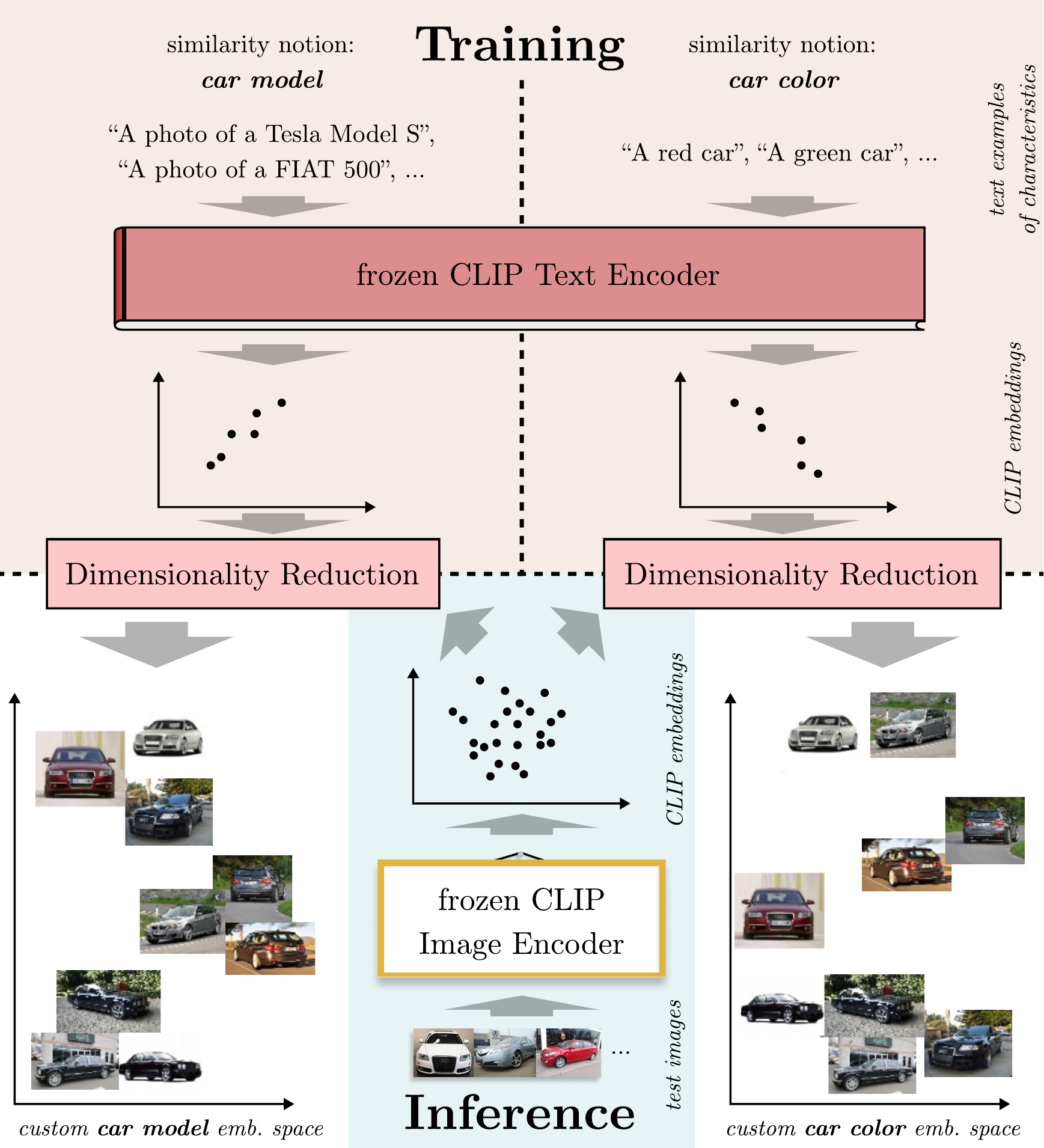}
    \caption{
        During \model's training (top half), different aspects of the desired similarity notion (e.g. car model or color) are collected in form of text prompts.
        CLIP's frozen text encoder embeds them and a dimensionality reduction method is learned to extract the dimensions that encode the similarity notion's aspects.
        During inference (bottom half), the trained dimensionality reduction is applied to CLIP encoded images to obtain custom image embeddings representing the desired similarity notion.
    }
    \label{fig:figure1}
\end{figure}

Deep Metric Learning (DML) is the task of training deep neural networks that map input items to a low-dimensional manifold such that similar items are represented by vectors close to each other~\cite{musgrave2020metric,kaya2019deep}.
In the usual DML setting, training examples are needed that let the model learn which image properties make an image pair (dis)similar.
For example, in the Cars196 dataset's setting~\cite{KrauseStarkDengFei-Fei_3DRR2013}, two images are deemed similar if they show the same car model.
Factors such as the car color, its orientation, or the image's environment should be suppressed by the embedding process.
Other datasets have different image properties to define when two images are similar.
We call this high-level interpretation of when two inputs are deemed similar a \textbf{similarity notion}.
For Cars196, the similarity notion is ``Two car images are similar if they show the same car model''.
During testing, the ability of the neural network to generalize this learned similarity notion to new unseen classes (e.g. new car models) is measured.

Often, people have different similarity notions depending on the task at hand or personal preference.
It is thus desirable to be able to quickly adapt to changing similarity notions.
However, large labeled training datasets are needed to train a model for a new similarity notion, which is time-consuming and tedious for users to create.
We thus aim for a zero-shot setting, where no training images and labels are needed.
We argue that users often can express the desired similarity notion using words, e.g., ``Two car images are similar if both cars have the same color''.
This is especially the case when there are categorical aspects with names that sort the images into disjoint classes.
More specifically, users can list a set of distinct aspects describing the similarity notion, e.g. ``a red car'', ``a green car'', ...
The use of language simplifies the process of expressing custom similarity notions, which alleviates the problem of collecting new labeled datasets.

As a first contribution of this work, we introduce a new task called \textbf{\longtask (\task)}:
Given a set of images $\mathcal{I}$ and a desired similarity notion $\mathcal{S}$ that is described using text $\mathcal{T}_\mathcal{S}$.
Train a Deep Metric Learning model using only the text input $\mathcal{T}_\mathcal{S}$ such that the resulting model can embed images $\mathcal{I} \rightarrow \mathbb{R}^n$ to $n$-dimensional embedding vectors, making image embeddings more similar if they are deemed similar regarding the similarity notion.
For optimization, no training images or labels are allowed (thus zero-shot).

Our second contribution is \model (\longmodel), a model for \task that uses a list of text prompts as input and learns a transformation that maps images to a vector space that reflects the desired similarity notion.
It utilizes the Contrastive Language-Image Pretraining (CLIP)~\cite{radford2021learning} model as a static general purpose feature extractor for images and texts.
We assume that CLIP embeddings for images and texts encode similar concepts in similar directions of the embedding space and that image descriptions can focus on certain properties.
For example, the text description ``a photo of a red car'' focuses on the car color and not on other features, such as the car's position, orientation, or environmental factors.

\Cref{fig:figure1} gives an overview of \model.
During training, CLIP's \textit{fixed} text encoder represents different characteristics of a desired similarity notion $\mathcal{S}$ as \num{512}-dimensional vectors, e.g. ``a red car'', ``a white car'', and so on to encode the car color.
We then extract the largest variations of these vectors in the embedding space by applying a dimensionality reduction method to these text representations, focusing on the changing aspects and abstracting away other non-related dimensions.
Learning the dimensionality reduction is fast and often, only a few dozen text prompts are needed.
Also, no training images or labels are used, only text prompts.

During inference, images are fed through CLIP's \textit{fixed} image encoder and the trained dimensionality reduction.
Assuming that CLIP's embeddings encode similar concepts in similar embedding space directions for both modalities, the resulting image representations are focused on the same dimensions as described by the text prompts.
Finally, lower-dimensional vectors can be used to find images similar to an image w.r.t. the desired similarity notion.

For our third contribution, we provide experimental evidence on five datasets and overall thirteen similarity notions, i.e. different properties by which the image embeddings should vary.
We show that \model consistently achieves better performance in retrieving similar images w.r.t. the desired similarity notion than strong baselines in the zero-shot setting and even approaches fully supervised baselines.
We also analyze the influence of changing embedding sizes, CLIP model sizes, and number of text prompts, and visualize the image regions \model focuses on to create image representations.
Our qualitative analysis shows that \model pays attention to pixels that are important in identifying the desired similarity notion.
Our code is publicly available.\footnote{\url{https://github.com/LSX-UniWue/InDiReCT}}

\section{Related Work}

Deep Metric Learning (DML) aims to learn neural networks that map input items to low-dimensional vectors such that similar items are close together in the embedding space~\cite{musgrave2020metric}.
In this work, we focus on images as items, which can be used for image retrieval~\cite{zhai2018classification}, face recognition~\cite{hu2014discriminative}, and image clustering~\cite{kaya2019deep}.
Usually, the model is trained on images, organized into classes, so binary similarity annotations are readily available for each pair of data points~\cite{kaya2019deep}.
Testing then uses a disjoint set of image classes to measure the model's generalization ability, but the data is semantically similar to the training data, e.g. Cars196~\cite{KrauseStarkDengFei-Fei_3DRR2013} only shows cars and face recognition datasets~\cite{hu2014discriminative} contain faces.

Studying DML generalization for images outside of the training domain has recently become popular~\cite{milbich2021characterizing,roth2020revisiting,milbich2020sharing,huai2019deep,hu2021generalization,xu2019zero}.
However, all proposed methods to improve the generalization performance to new datasets still use training images.
In our setting, no training images are allowed, but only text prompts to create an embedding space specifically tailored to a desired similarity notion.
For this, we use a fixed CLIP~\cite{radford2021learning} model to extract general purpose features.

The ability to rank possible text labels for an image using the cosine similarity of CLIP embeddings has been used in the original paper to perform zero-shot image classification~\cite{radford2021learning}.
For classification, the class names need to be known during inference while in \task, we create image embedding spaces reflecting the desired similarity notion.
Hence, the model needs to be able to handle images of unknown objects and characteristics, e.g. new car models.

Baldrati et al. use CLIP to alter fashion image embeddings using text prompts~\cite{baldrati2021conditioned}, e.g. the image of a black dress is combined with the text ``is red'' to find images of red dresses.
While exploiting similar properties of CLIP, we only use text prompts for training a transformation to focus on the desired similarity notion.
Image retrieval also uses joint text-image embeddings search for image contents using text~\cite{merkx2019language,chen2020uniter}.
We use text exclusively during training, not during inference.
To the best of our knowledge, no other work has a comparable task setting or method as our paper.

\section{Methodology}

We now introduce \model, our method for \longtask on images.
It makes use of a fixed CLIP~\cite{radford2021learning} as a general-purpose feature extractor for images and texts, which encodes similar concepts in both modalities to similar embedding directions.
CLIP consists of encoders for image and text.
It is pretrained on \num{400} million image-text pairs, optimized to embed both images and text such that embedding vectors of corresponding images and texts are more similar than different image-text pairs.
Similarity is measured using cosine similarity, i.e. the cosine of the angle between two vectors.
Due to the training task, CLIP learns to extract broad image features that can be correlated with/expressed by language.
Intuitively, we aim to learn a transformation that focuses on the most important features extracted by CLIP regarding the desired similarity notion.
\Cref{fig:figure1} shows \model's training and inference.

\subsection{Training}

In the training phase, $n$ different text prompts are created that describe certain characteristics of the desired similarity notion.
For example, if the target images show cars and we want to differentiate them by their color, we create a list of texts $\mathcal{T}_\mathcal{S}$ such as ``a red car'', ``a blue car'', ``a white car'', and so on.
The text prompts should only vary in the notion that we want to differentiate (here, the color descriptions).
Note that the aspects in the training text prompts are chosen independently from inference data, since inference labels are not known during training and we want to generalize to new aspects of the similarity notion as well.

When feeding all texts through CLIP's text encoder, the resulting $r$-dimensional vectors\footnote{$r = 512$ for CLIP's base model, but it is not limited to that number} $t_i \in \mathbb{R}^{1 \times r}$ for $i \in \{1, \dots, n\}$ vary in certain directions.
This is introduced by the change in aspects of the desired similarity notion in the text prompts.
Here, the variation of the vectors is only explained by the change of color names in the texts.

Due to CLIP encoding similar concepts to similar embedding dimensions, varying the same aspects in images and texts should result in embeddings that vary in similar directions.
Our goal is to find these directions using the text embeddings and suppress all other directions in the image embedding space, which are influenced by undesired factors.
Given the $n$ text representations $t_i \in \mathbb{R}^{1 \times r}$ for $i \in {1, \dots, n}$, we thus aim to identify the dimensions that vary the most in order to learn a transformation that retains these directions while reducing the embedding to $r'$ dimensions (similar to dimensionality reduction techniques such as PCA~\cite{wold1987principal}).
For this, we transform the text representations $t_i$ using a matrix $U \in \mathbb{R}^{r \times r'}$ and reconstruct them using $U^\top$.
We optimize $U$ with gradient descent to minimize reconstruction loss $L$:

\noindent
\begin{minipage}[c]{0.35\linewidth}
    \begin{align}
        &t^\text{norm}_i = \frac{t_i}{\lVert t_i \rVert} \label{eq:train1} \\
        &t'_i = \frac{t^\text{norm}_i U}{\lVert t^\text{norm}_i U \rVert}  \label{eq:train2}
    \end{align}
\end{minipage}
\begin{minipage}[c]{0.65\linewidth}
    \begin{align}
        &t^\text{recon}_i = \frac{t'_i U^\top}{\lVert t'_i U^\top \rVert}  \label{eq:train3} \\
        &L = \frac{1}{n} \sum_{i=1}^n \text{arccos}(t^\text{norm}_i {t^\text{recon}_i}^\top)  \label{eq:train4} \,.
    \end{align}
\end{minipage}

CLIP's use of cosine similarity as a similarity measure for embeddings disregards the length of all vectors, so we map input vectors $t_i$ and their reconstructions to a unit hypersphere (\Cref{eq:train1,eq:train3}).
Then we minimize the mean spherical distance (\Cref{eq:train4}) between the input and reconstructed vectors~\cite{kells1940plane}.
It is the distance between the vectors along the surface of the hypersphere, scaling linearly with the vector angle.
The training objective effectively minimizes the angles between the inputs and reconstructions.

In addition, the lower-dimensional embedding projections $t'_i$ are also mapped to a unit hypersphere (\Cref{eq:train2}).
This ensures that the reconstruction only uses the angles between the $r'$-dimensional vectors, keeping cosine similarity as a similarity measure in the lower-dimensional space, while preserving the varying directions of the text embeddings.

Since only up to a few hundred text prompts are used and only the matrix $U$ must be optimized, $L$ typically converges really fast.
The whole optimization process usually finishes in less than a minute on a common laptop's CPU, allowing \model to adapt to new similarity notions fast.

\subsection{Inference}

Given query and reference images, we feed them through CLIP's fixed image encoder and apply the learned transformation to map these embeddings $v_i \in \mathbb{R}^{1 \times r}$ ($i \in \{1, \dots, m\}$) to $r'$ dimensions on a unit hypersphere:

\noindent
\begin{minipage}[c]{0.4\linewidth}
    \begin{equation}
        v_i^\text{norm} = \frac{v_i}{\lVert v_i \rVert} \label{eq:inference1}
    \end{equation}
\end{minipage}
\begin{minipage}[c]{0.45\linewidth}
    \begin{equation}
        v'_i = \frac{v_i^\text{norm} U}{\lVert v_i^\text{norm} U \rVert} \label{eq:inference2} \,.
    \end{equation}
\end{minipage}

These vectors can be compared using the cosine/dot product similarity to find similar images w.r.t. the desired similarity notion.
Since the transformation learns to suppress dimensions that do not vary in the text prompts, these dimensions are also suppressed for images, e.g., a car \textit{model} dimension in the CLIP embedding space is suppressed when training with the similarity notion ``car color''.

\section{Experiments}

We now perform multiple experiments using \model and other baselines.
Since we are in a zero-shot learning setting, we have no access to labeled training images.
Hyperparameters cannot be tuned on a validation dataset, since labeled data is not allowed.
We thus define prompts and set hyperparameters based on commonly used values or educated guesses.
This resembles the real world scenario, where users do not have any training data at hand to verify and optimize their input to the system.

We implement \model using PyTorch~\cite{NEURIPS2019_9015} and sample the initial values of $U$ from $\mathcal{N}(0, 0.1)$.
We then optimize $U$ using Adam~\cite{kingma2014adam} with a learning rate of \num{0.01} until it does not improve the loss $L$ (\Cref{eq:train4}) for \num{100} consecutive iterations.
We reduce CLIP's vectors to \num{128} dimensions, which is a common embedding size for DML models~\cite{musgrave2020metric}.

We train models and compute image embeddings for each dataset and similarity notion.
We follow the standard evaluation setting of DML and measure the retrieval performance for these embeddings using the Mean Average Precision at R (MAP@R) and Precision at 1 (Prec@1)~\cite{musgrave2020metric}.
Results for other evaluation metrics are in the appendix.

\subsection{Datasets and Similarity Notions}

\begin{table*}[t]
\centering
\caption{Details on the datasets and similarity notions used for our experiments.}
\label{tab:datasets}
\resizebox{\linewidth}{!}{
    \begin{tabular}{@{}lp{3.5cm}p{2cm}p{9.5cm}p{7.2cm}@{}}
    \toprule
    \textbf{Dataset} & \textbf{Similarity Notion} & \textbf{Class Count} & \textbf{Prompt Template} & \textbf{Aspects (Count)} \\
    \midrule
    \multirow[t]{3}{*}{\textbf{Synthetic Cars~\cite{kobs2021different}}} & \textbf{Car Model} & \num{6} & ``a photo of a [car model]'' & Volvo S60, BMW X5 M, ... (\num{569}) \\
    \cmidrule{2-5}
    & \textbf{Car Color} & \num{18} & ``a [color name] car'' & orange, black, ... (\num{18}) \\
    \cmidrule{2-5}
    & \textbf{Background Color} & \num{18} & ``a car in front of a [color] background'' & orange, black, ... (\num{18}) \\
    \midrule
    \multirow[t]{3}{*}{\textbf{Cars196~\cite{KrauseStarkDengFei-Fei_3DRR2013}}} & \textbf{Car Model} & \num{98} & ``a photo of a [car model]'' & Volvo S60, BMW X5 M, ... (\num{569}) \\
    \cmidrule{2-5}
    & \textbf{Manufacturer} & \num{35} & ``a photo of a car produced by [manufacturer]'' & Tesla, BMW, ... (\num{46}) \\
    \cmidrule{2-5}
    & \textbf{Car Type} & \num{7} & ``a photo of a [car type]'' & convertible, SUV, ... (\num{7}) \\
    \midrule
    \textbf{CUB200~\cite{WahCUB_200_2011}} & \textbf{Bird Species} & \num{100} & ``a photo of a [bird species]'' & Black footed Albatross, Rusty Blackbird, ... (\num{100}) \\
    \midrule
    \multirow[t]{4}{*}{\textbf{DeepFashion~\cite{liu2016deepfashion}}} & \textbf{Clothing Category} & \num{50} & ``a photo of a person wearing a [clothing category]'' & anorak, turtleneck, ... (\num{50}) \\
    \cmidrule{2-5}
    & \textbf{Texture} & \num{7} & ``a photo of a person wearing clothes with a [texture type] texture'' & floral, striped, ... (\num{7}) \\
    \cmidrule{2-5}
    & \textbf{Fabric} & \num{6} & ``a photo of a person wearing clothes made out of [fabric type]'' & cotton, leather, ... (\num{6}) \\
    \cmidrule{2-5}
    & \textbf{Fit} & \num{3} & ``a photo of a person wearing clothes with a [fit type] fit'' & tight, loose, conventional (\num{3}) \\
    \midrule
    \multirow[t]{2}{*}{\textbf{Movie Posters~\cite{chu2017movie}}} & \textbf{Genre} & \num{25} & ``a poster of a [genre] movie'' & Comedy, Action, ... (\num{25}) \\
    \cmidrule{2-5}
    & \textbf{Production Country} & \num{69} & ``a poster of a movie produced in [country]'' & USA, India, ... (\num{69}) \\
    \bottomrule
    \end{tabular}
}
\end{table*}

We experiment with five datasets and overall thirteen similarity notions, which are listed in \Cref{tab:datasets}.
For each dataset, we define one to four similarity notions, e.g. the ``Car Model'' similarity notion of the Synthetic Cars~\cite{kobs2021different} and Cars196~\cite{KrauseStarkDengFei-Fei_3DRR2013} datasets can be expressed as ``Two car images are similar if they show the same car model''.
Other notions can be formulated accordingly.
Given a similarity notion, the datasets are split into different numbers of test classes (shown in the ``Class Count'' column), e.g. we use the \num{98} car models from Cars196's test dataset.
We create multiple text prompts for each similarity notion by collecting possible aspects and inserting them into a prompt template (listed in the corresponding columns).
The varying aspects are collected from different sources, such as an online car dealer website (``Car Model'' and ``Manufacturer''), the CSS2.1 color names (``Car Color'' and ``Background Color''), or the dataset's training data's labels (e.g. ``Bird Species'').
This promotes text prompts being possibly different from the test class labels, ensuring a realistic DML scenario, where train and test classes are commonly disjoint.
More details on datasets, similarity notions, and prompts are in the appendix.

\subsection{Baselines}

\model is the first method for \longtask, i.e. it can efficiently generate specialized embedding spaces for images based on the desired similarity notions.
Visualizations of embeddings produced by \model for similarity notions of the Cars196 dataset can be found in the appendix.
Since \model does not use any training images, it is not fair to compare it to fully supervised baselines.
However, we still contrast some of \model's results with fully supervised models and an Oracle baseline.
We use the following methods in our experiments to get a sense of how well \model performs.

\paragraph{Random Baseline}

For this baseline, we sample $r'$-dimensional embedding vectors for each image uniformly from the unit hypersphere~\cite{george1972choosing}.
This baseline indicates the performance lower bound for all methods.

\paragraph{CLIP~\cite{radford2021learning}}

This baseline feeds all images through CLIP's image encoder and uses the unmodified $r$-dimensional vectors as embeddings ($r = 512$).
Due to the broad set of features CLIP extracts, its performance should already be quite good.
However, since it does not focus on specific dimensions, \model is assumed to perform better while having fewer dimensions.
Even more so, CLIP cannot adapt its embeddings based on the desired similarity notion, i.e., it always yields the same embeddings for an image.
This limitation holds for all embedding methods that do not use additional data regarding the desired similarity notion.

\paragraph{Random Transformation}

\model optimizes a transformation that is applied to CLIP's image embeddings to achieve an embedding specialized towards a similarity notion expressed by text.
We evaluate how well the learning procedure of \model improves the performance by leaving $U$ as initialized for testing, i.e. sampled from $\mathcal{N}(0, 0.1)$.
We hypothesize that this baseline should, on average, be worse than both \model and the CLIP baseline.

\paragraph{Principal Component Analysis (PCA)~\cite{wold1987principal}}
\label{subsubsec:pca}

PCA is a popular dimensionality reduction technique which finds orthogonal directions that explain the largest variation in the data.
We test it as a possible alternative to our proposed method.
In contrast to our method, PCA solves for principal components analytically, requiring $r'$ to be strictly smaller than the number of input data points~\cite{scikitlearn}.
This is not satisfied for almost all scenarios in our experiments, since we only use a few text prompts while wanting to reduce CLIP's embeddings to a target size of \num{128} dimensions.
We thus can apply PCA only on the datasets that we collect more than \num{128} text prompts for, i.e. the ``Car Model'' similarity notion for the Synthetic Cars and Cars196 datasets.

\paragraph{Linear Autoencoder (LAE)}

The LAE is an alternative to PCA that provably spans the same subspace while being able to be trained using gradient descent~\cite{plaut2018principal}.
Formally, we optimize the weight matrices $W_1 \in \mathbb{R}^{r \times r'}$, $W_2 \in \mathbb{R}^{r' \times r}$ and bias vectors $b_1 \in \mathbb{R}^{1 \times r'}$, $b_2 \in \mathbb{R}^{1 \times r}$ with Adam (learning rate \num{0.01} and early stopping after \num{100} iterations) to minimize the loss function $L_\textit{\tiny LAE} = \sum_{i=1}^n \sum_{j} ((t_i^\text{norm})_j - (W_2(W_1 t_i^\text{norm} + b_1) + b_2)_j)^2$.
Image vectors are then transformed with $v'_i = W_1 v_i^\text{norm} + b_1$.

\paragraph{Nonlinear Autoencoder (AE)}

While PCA and LAE are linear models, we also test a more powerful nonlinear Autoencoder, which consists of a two-layer encoder and decoder with \num{512} hidden units and leaky ReLU activation functions~\cite{maas2013rectifier}.
We use the same loss function and hyperparameters as for LAE, but add a weight decay of \num{e-2} to alleviate overfitting on the few text prompts.

\paragraph{Oracle}
\label{subsubsec:oracle}

\model uses only text prompts to optimize the transformation matrix $U$ that maps CLIP embeddings to a more specialized, lower-dimensional unit hypersphere.
To estimate how well \model could theoretically perform, we employ an Oracle that optimizes $U$ directly on \textit{test images and their labels}.
For this, we use the common DML loss function Normalized Softmax Loss~\cite{zhai2018classification}.
We first compute unit-length image embeddings $v'_i$ as in \Cref{eq:inference1,eq:inference2} and then optimize the transformation matrix $U$ to minimize the loss function $L_\textit{\tiny Oracle} = \frac{1}{m} \sum_{i=1}^{m} - \log \left( \frac{\exp(v'_i c_{l_i}^\top)}{\sum_{j} \exp(v'_i c_j^\top))} \right)$, where $m$ is the number of test images and $c_{l_i} \in \mathbb{R}^{1 \times r'}$ with $\lVert c_{l_i} \rVert = 1$ is the prototype vector of the class for the label of the $i$th image $l_i$, which is optimized jointly with $U$ using Adam (learning rate \num{0.01}, early stopping with patience \num{100}).

Note that in \longtask, neither images nor their labels are available for training.
We use this baseline method in order to provide a \textit{very optimistic} estimate of what performance \model could achieve given perfect information.
The Normalized Softmax Loss is a classification-based training objective, so image embeddings are processed independently.
Thus, the loss does not optimize for the best nearest neighbor performance, i.e. Precision@1.
To compare the Oracle baseline to other models, we thus primarily use MAP@R.

Low (high) Oracle performance can be used to identify similarity notions that cannot (can) be reliably represented using \model since they are not captured (are captured) in the CLIP embeddings.
If \model's performs substantially worse than the Oracle, it means that the text prompts were not capable of capturing the desired similarity notion.

\section{Results}

\begin{table*}[t]
\centering
\caption{Results for our experiments. All values are given in percent, best in bold.}
\label{tab:results}
\resizebox{\linewidth}{!}{
    \begin{tabular}{@{}lrrccccccc|c@{}}
    \toprule
     &  &  & \textbf{Random} & \textbf{CLIP (512-dim.)} & \textbf{\model} & \textbf{Rand. trans.} & \textbf{PCA} & \textbf{LAE} & \textbf{AE} & \textbf{Oracle} \\ 
    \midrule
    \multirow{6}{*}{\textbf{Synthetic Cars}} & \multirow{2}{*}{\textbf{Car Model}} & \textbf{MAP@R} & 3.3 $\pm$ 0.1 & 43.5 & \textbf{57.4 $\pm$ 0.2} & 39.1 $\pm$ 1.6 & 56.2 $\pm$ 0.1 & 52.5 $\pm$ 0.5 & 39.5 $\pm$ 4.4 & 100 $\pm$ 0.0 \\
     &  & \textbf{Prec@1} & 17.5 $\pm$ 0.9 & 95.4 & 96.4 $\pm$ 0.0 & 93.4 $\pm$ 0.5 & \textbf{96.6 $\pm$ 0.1} & 95.9 $\pm$ 0.5 & 88.7 $\pm$ 3.6 & 100 $\pm$ 0.0 \\ 
    \cmidrule{2-11}
     & \multirow{2}{*}{\textbf{Car Color}} & \textbf{MAP@R} & 5.0 $\pm$ 0.1 & 6.2 & \textbf{9.1 $\pm$ 0.1} & 6.1 $\pm$ 0.1 & --- & 7.3 $\pm$ 0.2 & 8.6 $\pm$ 0.4 & 57.9 $\pm$ 0.9 \\
     &  & \textbf{Prec@1} & 17.5 $\pm$ 0.8 & 27.6 & \textbf{31.4 $\pm$ 0.5} & 26.3 $\pm$ 1.3 & --- & 29.4 $\pm$ 0.9 & 30.2 $\pm$ 1.3 & 79.3 $\pm$ 0.8 \\ 
    \cmidrule{2-11}
     & \multirow{2}{*}{\textbf{Background Color}} & \textbf{MAP@R} & 5.4 $\pm$ 0.0 & 6.2 & \textbf{7.1 $\pm$ 0.0} & 6.1 $\pm$ 0.2 & --- & 6.3 $\pm$ 0.2 & 6.1 $\pm$ 0.2 & 74.0 $\pm$ 0.9 \\
     &  & \textbf{Prec@1} & 19.4 $\pm$ 1.1 & 27.0 & \textbf{28.3 $\pm$ 0.3} & 26.6 $\pm$ 1.1 & --- & \textbf{28.3 $\pm$ 0.7} & 21.6 $\pm$ 1.3 & 88.0 $\pm$ 0.4 \\ 
    \midrule
    \multirow{6}{*}{\textbf{Cars196}} & \multirow{2}{*}{\textbf{Car Model}} & \textbf{MAP@R} & 0.1 $\pm$ 0.0 & 23.5 & 37.4 $\pm$ 0.0 & 19.2 $\pm$ 0.3 & \textbf{37.5 $\pm$ 0.1} & 33.2 $\pm$ 0.2 & 20.0 $\pm$ 5.8 & 41.8 $\pm$ 0.0 \\
     &  & \textbf{Prec@1} & 1.1 $\pm$ 0.1 & 78.0 & \textbf{84.4 $\pm$ 0.1} & 72.9 $\pm$ 0.5 & 84.2 $\pm$ 0.1 & 82.4 $\pm$ 0.2 & 63.8 $\pm$ 8.1 & \textit{76.6 $\pm$ 0.1} \\
    \cmidrule{2-11}
     & \multirow{2}{*}{\textbf{Manufacturer}} & \textbf{MAP@R} & 0.5 $\pm$ 0.0 & 24.4 & \textbf{33.6 $\pm$ 0.1} & 21.2 $\pm$ 0.4 & --- & 24.2 $\pm$ 0.4 & 18.0 $\pm$ 2.2 & 51.4 $\pm$ 0.0 \\
     &  & \textbf{Prec@1} & 5.4 $\pm$ 0.3 & 89.0 & \textbf{90.5 $\pm$ 0.1} & 84.7 $\pm$ 0.8 & --- & 85.5 $\pm$ 0.3 & 63.1 $\pm$ 3.9 & \textit{84.0 $\pm$ 0.1} \\ 
    \cmidrule{2-11}
     & \multirow{2}{*}{\textbf{Car Type}} & \textbf{MAP@R} & 3.5 $\pm$ 0.0 & 25.1 & \textbf{36.1 $\pm$ 0.3} & 22.1 $\pm$ 0.8 & --- & 27.7 $\pm$ 0.6 & 24.4 $\pm$ 1.6 & 73.8 $\pm$ 0.0 \\
     &  & \textbf{Prec@1} & 17.3 $\pm$ 0.4 & \textbf{91.1} & 90.7 $\pm$ 0.2 & 88.3 $\pm$ 0.5 & --- & 89.1 $\pm$ 0.4 & 63.2 $\pm$ 3.1 & \textit{89.1 $\pm$ 0.0} \\ 
    \midrule
    \multirow{2}{*}{\textbf{CUB200}} & \multirow{2}{*}{\textbf{Bird Species}} & \textbf{MAP@R} & 0.1 $\pm$ 0.0 & 18.0 & \textbf{26.5 $\pm$ 0.0} & 15.2 $\pm$ 0.3 & --- & 18.8 $\pm$ 0.2 & 15.1 $\pm$ 1.9 & 34.1 $\pm$ 0.0 \\
     &  & \textbf{Prec@1} & 1.2 $\pm$ 0.1 & 58.2 & \textbf{65.3 $\pm$ 0.1} & 52.6 $\pm$ 0.3 & --- & 58.1 $\pm$ 0.5 & 44.4 $\pm$ 3.6 & \textit{65.3 $\pm$ 0.2} \\ 
    \midrule
    \multirow{8}{*}{\textbf{DeepFashion}} & \multirow{2}{*}{\textbf{Clothing Category}} & \textbf{MAP@R} & 2.3 $\pm$ 0.0 & 12.5 & \textbf{18.7 $\pm$ 0.1} & 11.3 $\pm$ 0.4 & --- & 13.3 $\pm$ 0.3 & 16.9 $\pm$ 1.8 & 32.2 $\pm$ 0.1 \\
     &  & \textbf{Prec@1} & 11.1 $\pm$ 0.4 & 45.2 & \textbf{50.9 $\pm$ 0.2} & 43.0 $\pm$ 0.6 & --- & 45.5 $\pm$ 0.5 & 44.5 $\pm$ 2.4 & 55.8 $\pm$ 0.6 \\ 
    \cmidrule{2-11}
     & \multirow{2}{*}{\textbf{Texture}} & \textbf{MAP@R} & 11.8 $\pm$ 0.0 & 18.7 & \textbf{33.0 $\pm$ 0.4} & 11.2 $\pm$ 0.4 & --- & 22.2 $\pm$ 0.5 & 16.3 $\pm$ 0.7 & 66.1 $\pm$ 0.1 \\
     &  & \textbf{Prec@1} & 29.6 $\pm$ 0.7 & 60.2 & \textbf{66.8 $\pm$ 0.3} & 43.3 $\pm$ 0.5 & --- & 61.2 $\pm$ 0.7 & 43.8 $\pm$ 1.7 & 80.6 $\pm$ 0.3 \\ 
    \cmidrule{2-11}
     & \multirow{2}{*}{\textbf{Fabric}} & \textbf{MAP@R} & 32.4 $\pm$ 0.0 & 34.0 & \textbf{37.7 $\pm$ 0.2} & 10.8 $\pm$ 0.3 & --- & 35.6 $\pm$ 0.3 & 17.2 $\pm$ 0.6 & 64.2 $\pm$ 0.3 \\
     &  & \textbf{Prec@1} & 49.4 $\pm$ 0.6 & 64.5 & \textbf{66.1 $\pm$ 0.6} & 42.6 $\pm$ 0.7 & --- & 65.1 $\pm$ 0.6 & 44.7 $\pm$ 1.9 & 77.8 $\pm$ 0.4 \\ 
    \cmidrule{2-11}
     & \multirow{2}{*}{\textbf{Fit}} & \textbf{MAP@R} & 51.8 $\pm$ 0.0 & 53.3 & \textbf{53.9 $\pm$ 0.4} & 11.1 $\pm$ 1.0 & --- & 53.4 $\pm$ 0.3 & 16.1 $\pm$ 1.8 & 82.0 $\pm$ 0.1 \\
     &  & \textbf{Prec@1} & 66.6 $\pm$ 0.6 & \textbf{77.1} & 76.5 $\pm$ 0.4 & 43.1 $\pm$ 0.5 & --- & 76.7 $\pm$ 0.7 & 42.9 $\pm$ 1.9 & 87.8 $\pm$ 0.6 \\ 
    \midrule
    \multirow{4}{*}{\textbf{Movie Posters}} & \multirow{2}{*}{\textbf{Genre}} & \textbf{MAP@R} & 4.1 $\pm$ 0.0 & 11.4 & \textbf{14.9 $\pm$ 0.0} & 9.1 $\pm$ 0.3 & --- & 8.4 $\pm$ 0.1 & 9.8 $\pm$ 2.4 & 19.6 $\pm$ 0.1 \\
     &  & \textbf{Prec@1} & 17.5 $\pm$ 0.4 & 41.8 & \textbf{44.0 $\pm$ 0.2} & 38.1 $\pm$ 0.7 & --- & 36.6 $\pm$ 0.4 & 33.3 $\pm$ 3.0 & \textit{43.2 $\pm$ 0.7} \\ 
    \cmidrule{2-11}
     & \multirow{2}{*}{\textbf{Production Country}} & \textbf{MAP@R} & 44.6 $\pm$ 0.0 & 49.3 & \textbf{51.3 $\pm$ 0.1} & 48.9 $\pm$ 0.4 & --- & 47.7 $\pm$ 0.2 & 49.4 $\pm$ 0.7 & 58.1 $\pm$ 0.0 \\
     &  & \textbf{Prec@1} & 59.2 $\pm$ 0.5 & 69.3 & \textbf{69.8 $\pm$ 0.3} & 67.9 $\pm$ 0.7 & --- & 68.1 $\pm$ 0.3 & 64.9 $\pm$ 0.7 & 71.8 $\pm$ 0.3 \\
    \bottomrule
    \end{tabular}
}
\end{table*}

We report the mean and standard deviation of the evaluation metrics over five runs in \Cref{tab:results}.
The CLIP baseline typically achieves substantially better results than the random baseline.
Since the embeddings stay the same in each run, its performance does have a standard deviation of zero and is omitted for brevity.
Despite the fact that the CLIP baseline uses four times larger embedding vectors, \model almost always performs better than CLIP and achieves the best performance in most datasets and similarity notions.
Depending on the dataset and similarity notion, \model can improve CLIP's MAP@R score by up to \num{14} percentage points.
Switching the learned matrix to a random transformation matrix in \model usually performs worse than CLIP.
As described in \Cref{subsubsec:pca}, PCA is only applicable to two datasets and similarity notions.
There, \model and PCA perform similarly.
Training a Linear Autoencoder (LAE) on the text embeddings usually improves the CLIP baseline, but does not achieve better performance than \model.
Applying a more complex nonlinear Autoencoder performs oftentimes worse than the CLIP baseline and also shows substantially larger standard deviations, which might be due to the model not handling the few datapoints well.
These results show that choosing a suitable dimensionality reduction technique can improve performance and opens up new research directions.
In general, \model learns a useful embedding function by using text prompts that describe different aspects of the desired similarity notion.

The Oracle baseline is optimized directly on the image dataset and their labels.
Despite all this, \model matches or exceeds the Prec@1 performance of the Oracle baseline for Cars196, CUB200, and the ``Genre'' similarity notion for the Movie Posters dataset.
As discussed in \Cref{subsubsec:oracle}, this might be due to the classification-based nature of the Normalized Softmax Loss.
For MAP@R, the Oracle is the best model for all datasets and similarity notions.

Even though the comparison is not fair, we contrast \model's performance with state of the art models from the literature that train on a large labeled training dataset regarding the desired similarity notion.
Note that only Cars196's ``Car Model'' and CUB200's ``Bird Species'' similarity notions have been used in the literature in a DML setting, so we only compare to them.
Jun et al.~\cite{jun2019combination} achieve Prec@1 of \num{94.8} and \num{79.2} for Cars196 and CUB200, respectively~\cite{rendle2019evaluation}, which outperform \model by ten to fourteen percentage points.
However, the trained models output \num{1536}-dimensional vectors, more than ten times the embedding dimensions we use in our experiments.
For embeddings of dimensions \num{128}, Jun et al. achieve \num{90.1} (Cars196) and \num{67.6} (CUB200) Prec@1, which is only approximately six and two percentage points better than \model.
These results show that despite not using any training images, \model can show strong performance even compared to fully supervised methods.

\section{Analysis}

\paragraph{What does \model attend to in the input?}

\begin{figure}[t]
    \centering

    \includegraphics[width=0.24\linewidth]{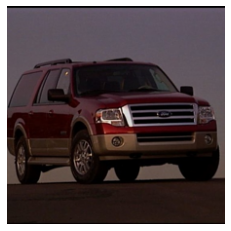}
    \hfill
    \includegraphics[width=0.24\linewidth]{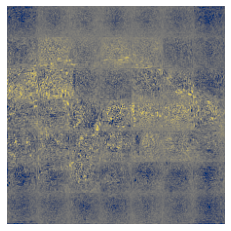}
    \hfill
    \includegraphics[width=0.24\linewidth]{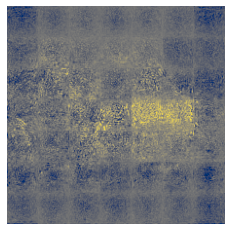}
    \hfill
    \includegraphics[width=0.24\linewidth]{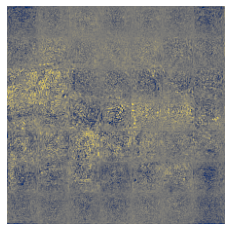}
    
    \includegraphics[width=0.24\linewidth]{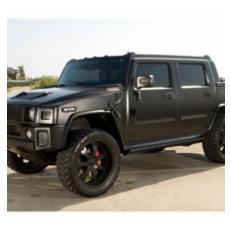}
    \hfill
    \includegraphics[width=0.24\linewidth]{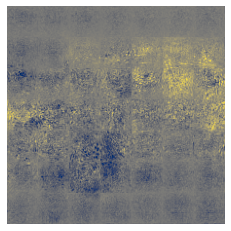}
    \hfill
    \includegraphics[width=0.24\linewidth]{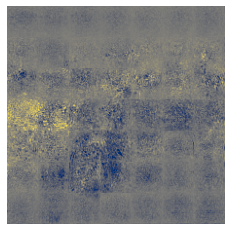}
    \hfill
    \includegraphics[width=0.24\linewidth]{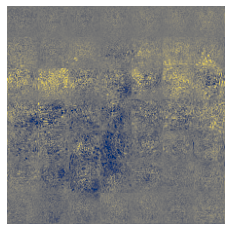}

    \vspace{-2mm}
    \begin{tabularx}{\linewidth}{YYYY}
        \textbf{\scriptsize \centering Image} &
        \textbf{\scriptsize \centering Car Model} &
        \textbf{\scriptsize \centering Manufacturer} &
        \textbf{\scriptsize \centering Car Type}
    \end{tabularx}
    \vspace{-3mm}
    
    \caption{Example images from the Cars196 dataset and the saliency map differences between each similarity notion and CLIP.
    \model focuses more on yellow regions, CLIP more on blue regions.
    The patch patterns in the images are due to the patch creation of CLIP's Vision Transformer~\cite{dosovitskiy2020image}. More examples in the appendix.}
    \label{fig:explainer_analysis}
\end{figure}

We want to visualize the image regions that are used by \model to output a certain embedding.
Due to the positive experimental results, we assume that, for a given similarity notion, \model attends to subjectively more useful regions than CLIP.
We thus compute saliency maps using the method introduced by Kobs et al.~\cite{kobs2021different} and subtract \model's saliency maps from CLIP's saliency maps to qualitatively showcase the difference between both methods.

We choose Cars196 and its similarity notions and hypothesize that \model pays more attention to regions that represent the desired similarity notion than CLIP.
In order to increase the chance of obtaining visible differences in the saliency maps, we reduce the number of embedding dimensions for \model to two, thus only extracting the most important features to embed the given images.
\Cref{fig:explainer_analysis} shows two example images (more in the appendix).
Yellow areas indicate image regions \model pays more attention to than CLIP, while CLIP focuses more on blue regions.
Grey areas show similarly strong saliency.

Compared to CLIP, \model focuses more on the area of the car when using the ``Car Model'' similarity notion, which is useful for the task.
Interestingly, for ``Manufacturer'', \model mostly uses the front of the car, where the manufacturer's logo is usually found.
Additionally, the design of the radiator grill and headlights is often relatively unique to manufacturers.
For the ``Car Type'' similarity notion, \model focuses more on the back of the car, as car types such as ``convertible'', ``van'', or ``sedan'' differ mainly in terms of trunk and roof design.

\paragraph{Do other embedding sizes perform differently?}

\begin{figure*}[t]
    \centering
    \includegraphics[width=\linewidth]{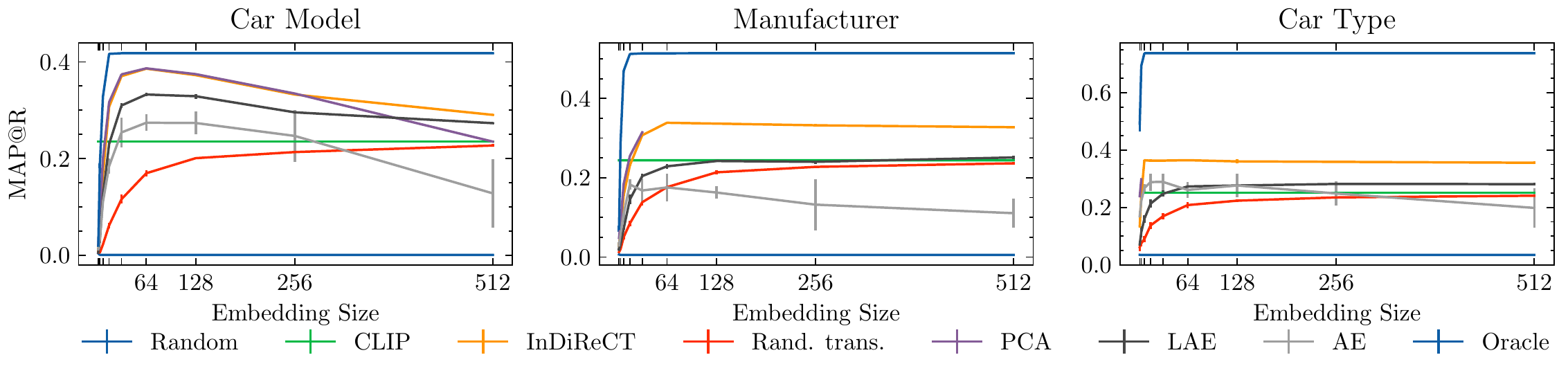}

    \caption{On Cars196, \model outperforms other zero-shot models for embedding sizes \num{16} and up, while it peaks at \num{64} dimensions.}
    \label{fig:size_analysis}
\end{figure*}

While our experiments set the embedding size arbitrarily to \num{128}, we now measure the performance on the Cars196 dataset with varying target embedding dimensions $r' \in \{2, 4, 8, \dots, 256, 512\}$.
We plot the MAP@R mean and standard deviation over five runs for all methods and all similarity notions in \Cref{fig:size_analysis}.
CLIP with its fixed \num{512} dimensions is plotted as a reference line.

\model matches or exceeds CLIP's performance when using at least \num{16} embedding dimensions and peaks at \num{64} dimensions for all three similarity notions.
The learned transformation presumably selects, combines, and weights CLIP's embedding dimensions such that \model even outperforms CLIP for \num{512} dimensions.

\paragraph{Do larger CLIP models improve performance?}

\begin{figure*}[t]
    \centering
    \includegraphics[width=\linewidth]{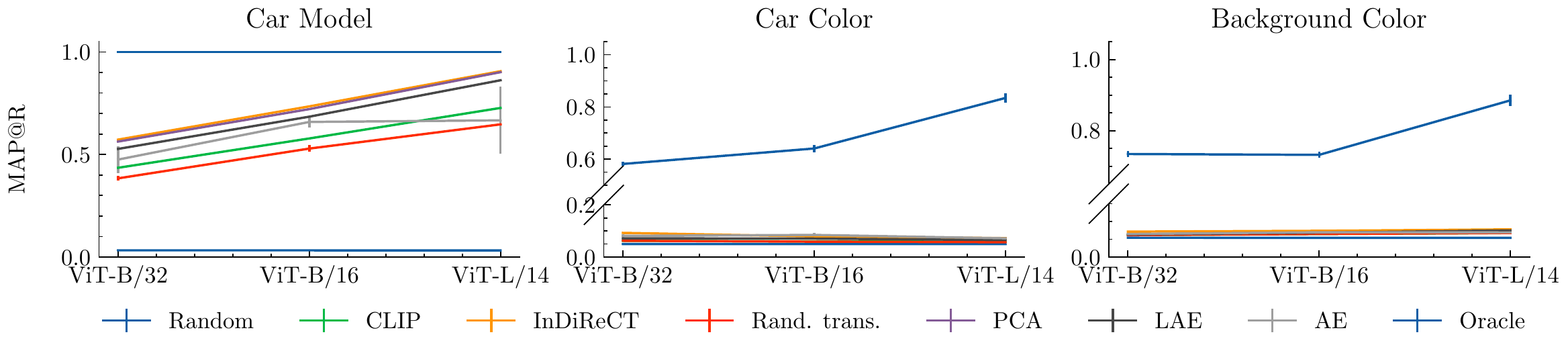}

    \caption{Larger CLIP models improve performance for the ``Car Model'' but not for color similarity notions on Synthetic Cars.}
    \label{fig:vit_size_analysis}
\end{figure*}

For our experiments, we use the CLIP model ``ViT-B/32''~\cite{radford2021learning}, i.e. a Vision Transformer~\cite{dosovitskiy2020image} with \num{12} layers and input patches of size $32 \times 32$ pixels.
We now test larger CLIP models as feature extractors in \model with CLIP's ``ViT-B/16'' and ``ViT-L/14'' versions, which change the input patches to $16 \times 16$ and $14 \times 14$ pixels, respectively, while ``ViT-L/14'' also doubles the transformer layers.
Besides other parameters, ``ViT-L/14'' also increases CLIP's outputs from \num{512}- to \num{768}-dimensional vectors.

We test all three ViT sizes to see if larger CLIP versions lead to better performance~\cite{radford2021learning}.
The ``Synthetic Cars'' dataset with its similarity notions is used, since the performance of \model is quite good for ``Car Model'', but bad for ``Car Color'' and ``Background Color'', compared to the Oracle baseline.
With this analysis, we can investigate whether larger models can improve performance for these similarity notions.
We use \num{128} embedding dimensions.

\Cref{fig:vit_size_analysis} shows that the performance of the Oracle baseline increases with larger models, which means that the model extracts more useful features that could potentially be picked up by \model.
For the ``Car Model'' similarity notion, this also translates to better performance of \model and CLIP in general.
On the other two similarity notions, however, we cannot find any performance improvements.
Since the Oracle baseline improves, we can conclude that the text prompts used to train \model lead to a focus on suboptimal features for these similarity notions.
Other text prompts might increase performance.

\paragraph{Do more text prompts improve performance?}

\begin{figure}[t]
    \centering
    \includegraphics[width=\linewidth]{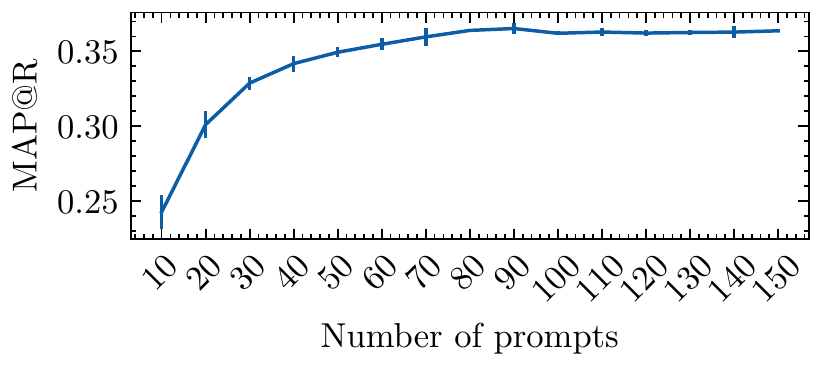}
    
    \caption{Performance of \model for different number of training prompts. We sample different car model names for each run.}
    \label{fig:prompt_size_analysis}
\end{figure}

Our final analysis takes a closer look at how the performance of \model changes if we use different numbers of prompts for our experiments.
We use the Cars196 dataset and focus on the ``Car Model'' similarity notion.
Originally, we use \num{569} different car model names from an online car dealer as a basis for the text prompts (``a photo of a [car model name]'').
We now sample differently sized sets from these car model names and run our experiment five times with different samples.
\Cref{fig:prompt_size_analysis} shows the means and standard deviations for sizes $\{10, 20, \dots, 150\}$.
The performance increases with larger sample sizes and converges at around \num{90} prompts to the performance we observe in our main experiments.
This behavior is expected, since the learned transformation is able to better capture the important dimensions in the text embeddings when more prompts are used.
For fewer prompts, \model can almost perfectly reconstruct the text embeddings, thus is not forced to select the important dimensions.
\Cref{fig:prompt_size_analysis} also shows that with larger prompt sets, the standard deviation of performance tends to decrease.
Overall, we can observe that more (useful) text prompts should stabilize and improve performance for \model.

\section{Discussion}

Using natural language, the proposed \task setting offers a simple interface for adapting item retrieval systems to the desired similarity notion.
This adaption is not achievable using raw CLIP embeddings or other self-/unsupervised methods.
For \model, it is not necessary to collect and annotate example images, which is time-consuming and tedious.
Expressing the desired similarity notion using text prompts is certainly simpler, but limits its application to similarity notions with categorical aspects.
However, this is a limitation that also holds for popular proxy-based DML loss functions such as Normalized Softmax Loss~\cite{zhai2018classification} or ProxyNCA~\cite{movshovitz2017no}, i.e., loss functions that use class prototype vectors.
It should also be noted that the quality of text prompts might vary significantly.
In our experiments, we comply with the zero-shot setting by choosing plausible prompt templates without validating them on the data.
Overall, we achieve good performance across datasets and similarity notions.
However, as already shown for prompt engineering~\cite{radford2021learning}, there might be prompts that work substantially better.
Often, exploiting the peculiarities of the dataset CLIP has been trained on helps.
For example, instead of using single words as text prompts, short sentences usually work better~\cite{radford2021learning}.
Therefore, it is recommended to test different text prompts when applied in real-world scenarios.
Also, tuning the number of embedding dimensions is not straightforward without validation data, leading to suboptimal performance when using \num{128}- instead of \num{64}-dimensional vectors for the Cars196 dataset, as shown in our analysis.

Since we use CLIP as a fixed feature extractor, we need to rely on the usefulness of its embeddings.
If CLIP does not extract properties from images and texts related to a desired similarity notion, \model cannot show its full potential.
We have shown that \model mostly outperforms CLIP, so the text prompts help to focus on the desired similarity notion.
Given the Oracle results, however, some datasets and similarity notions (e.g. Synthetic Cars' color notions) could potentially work better.
In some cases, larger CLIP models can improve the performance as shown in our analysis.

Since we use pretrained CLIP embeddings and only a handful of text prompts, training the dimensionality reduction is fast.
It also allows us to precompute CLIP embeddings for a whole image database and adaptively transform them with a trained dimensionality reduction.
The disadvantage of this is that, for each search, the transformation matrix must be applied to all vectors in the image collection.
Potentially, existing vector search databases~\cite{2021milvus,johnson2019billion} can efficiently incorporate the transformation to retrieve relevant images.

\section{Conclusion}

In this paper, we have introduced \longtask~(\task), a setting where no training data and labels but only texts are allowed to guide a Deep Metric Learning model for a given similarity notion.
Our proposed model \model is based on fixed CLIP embeddings of text prompts describing the varying aspects of a given similarity notion.
We have shown that \model outperforms strong baselines and approaches fully supervised methods.
Our analyses show that \model focuses on image regions that are subjectively important for the desired similarity notion.
We have also investigated the influence of different hyperparameters on the model performance.

Due to its simple design and fast training, \model can be useful for users to customize the similarity notion of item retrieval systems.
The need to define multiple prompts based on the changing aspects of a similarity notion could be facilitated, e.g. by directly learning the transformation from sentences such as ``Two car images are similar if both cars are the same model''.
Automatic selection of hyperparameters and developing methods for \task on other modalities, e.g. audio or texts, are also interesting research directions.

{\small
\bibliographystyle{ieee_fullname}
\bibliography{egbib}
}

\end{document}


\maketitle

\section{Full Results}

\begin{table}[t]
\centering
\caption{Results for our experiments on five datasets and thirteen similarity notions.}
\label{tab:results}
\resizebox*{!}{0.95\textheight}{
    \begin{tabular}{@{}lrrccccccc|c@{}}
    \toprule
    & & {} &            \textbf{Random} &        \textbf{CLIP} & \textbf{\model} & \textbf{Rand. trans.} & \textbf{PCA} & \textbf{LAE} & \textbf{AE} & \textbf{Oracle} \\
    \midrule
    \multirow{21}{*}{\textbf{Synthetic Cars}} & \multirow{7}{*}{\textbf{Car Model}} & \textbf{MAP@R} &   0.033 $\pm$ 0.001 &  0.435 &       \textbf{0.574 $\pm$ 0.002} &       0.391 $\pm$ 0.016 & 0.562 $\pm$ 0.001 & 0.526 $\pm$ 0.005 & 0.395 $\pm$ 0.044 & 1.000 $\pm$ 0.000 \\
    &&\textbf{Prec@1} &   0.175 $\pm$ 0.009 &  0.954 &       0.964 $\pm$ 0.000 &       0.934 $\pm$ 0.005 & \textbf{0.966 $\pm$ 0.001} & 0.959 $\pm$ 0.005 & 0.887 $\pm$ 0.036 & 1.000 $\pm$ 0.000 \\
    &&\textbf{R-Prec} &   0.167 $\pm$ 0.001 &  0.548 &       \textbf{0.662 $\pm$ 0.001} &       0.510 $\pm$ 0.014 & 0.653 $\pm$ 0.001 & 0.624 $\pm$ 0.004 & 0.517 $\pm$ 0.036 & 1.000 $\pm$ 0.000 \\
    
    &&\textbf{AMI} &  -0.000 $\pm$ 0.002 &  0.623 &       \textbf{0.737 $\pm$ 0.006} &       0.559 $\pm$ 0.086 & 0.730 $\pm$ 0.004 & 0.713 $\pm$ 0.010 & 0.539 $\pm$ 0.060 & 0.896 $\pm$ 0.000 \\
    &&\textbf{NMI} &   0.007 $\pm$ 0.002 &  0.626 &       \textbf{0.738 $\pm$ 0.006} &       0.562 $\pm$ 0.085 & 0.732 $\pm$ 0.004 & 0.715 $\pm$ 0.010 & 0.542 $\pm$ 0.059 & 0.897 $\pm$ 0.000 \\
    &&\textbf{MAP} &   0.172 $\pm$ 0.001 &  0.591 &       \textbf{0.716 $\pm$ 0.002} &       0.546 $\pm$ 0.016 & 0.706 $\pm$ 0.001 & 0.675 $\pm$ 0.005 & 0.551 $\pm$ 0.041 & 1.000 $\pm$ 0.000 \\
    &&\textbf{MRR} &   0.367 $\pm$ 0.009 &  0.974 &       0.980 $\pm$ 0.000 &       0.961 $\pm$ 0.003 & \textbf{0.981 $\pm$ 0.000} & 0.977 $\pm$ 0.003 & 0.928 $\pm$ 0.024 & 1.000 $\pm$ 0.000 \\
    
    \cmidrule{2-11}
    & \multirow{7}{*}{\textbf{Car Color}} & \textbf{MAP@R} &   0.050 $\pm$ 0.001 &  0.062 &       \textbf{0.091 $\pm$ 0.001} &       0.061 $\pm$ 0.001 & --- & 0.073 $\pm$ 0.002 & 0.086 $\pm$ 0.004 & 0.579 $\pm$ 0.009 \\
    &&\textbf{Prec@1} &   0.175 $\pm$ 0.008 &  0.276 &       \textbf{0.314 $\pm$ 0.005} &       0.263 $\pm$ 0.013 & --- & 0.294 $\pm$ 0.009 & 0.302 $\pm$ 0.013 & 0.793 $\pm$ 0.008 \\
    &&\textbf{R-Prec} &   0.174 $\pm$ 0.001 &  0.198 &       \textbf{0.250 $\pm$ 0.002} &       0.196 $\pm$ 0.002 & --- & 0.218 $\pm$ 0.001 & 0.238 $\pm$ 0.004 & 0.676 $\pm$ 0.007 \\
    
    &&\textbf{AMI} &  -0.003 $\pm$ 0.004 &  0.029 &       \textbf{0.170 $\pm$ 0.003} &       0.027 $\pm$ 0.008 & --- & 0.073 $\pm$ 0.002 & 0.156 $\pm$ 0.016 & 0.629 $\pm$ 0.009 \\
    &&\textbf{NMI} &   0.058 $\pm$ 0.004 &  0.088 &       \textbf{0.220 $\pm$ 0.003} &       0.086 $\pm$ 0.008 & --- & 0.129 $\pm$ 0.002 & 0.207 $\pm$ 0.015 & 0.652 $\pm$ 0.008 \\
    &&\textbf{MAP} &   0.179 $\pm$ 0.001 &  0.197 &       \textbf{0.247 $\pm$ 0.002} &       0.196 $\pm$ 0.002 & --- & 0.213 $\pm$ 0.001 & 0.236 $\pm$ 0.005 & 0.712 $\pm$ 0.007 \\
    &&\textbf{MRR} &   0.336 $\pm$ 0.007 &  0.451 &       \textbf{0.493 $\pm$ 0.003} &       0.439 $\pm$ 0.014 & --- & 0.474 $\pm$ 0.008 & 0.477 $\pm$ 0.011 & 0.859 $\pm$ 0.006 \\
    
    \cmidrule{2-11}
    & \multirow{7}{*}{\textbf{\makecell[r]{Background\\Color}}} & \textbf{MAP@R} &  0.054 $\pm$ 0.000 &  0.062 &       \textbf{0.071 $\pm$ 0.000} &       0.061 $\pm$ 0.002 & --- & 0.063 $\pm$ 0.002 & 0.061 $\pm$ 0.002 & 0.740 $\pm$ 0.009 \\
    &&\textbf{Prec@1} &  0.194 $\pm$ 0.011 &  0.270 &       \textbf{0.283 $\pm$ 0.003} &       0.266 $\pm$ 0.011 & --- & \textbf{0.283 $\pm$ 0.007} & 0.216 $\pm$ 0.013 & 0.880 $\pm$ 0.004 \\
    &&\textbf{R-Prec} &  0.183 $\pm$ 0.001 &  0.200 &       \textbf{0.218 $\pm$ 0.001} &       0.199 $\pm$ 0.003 & --- & 0.204 $\pm$ 0.003 & 0.197 $\pm$ 0.004 & 0.805 $\pm$ 0.007 \\
    
    &&\textbf{AMI} &  0.004 $\pm$ 0.003 &  0.017 &       \textbf{0.089 $\pm$ 0.003} &       0.025 $\pm$ 0.012 & --- & 0.039 $\pm$ 0.006 & 0.048 $\pm$ 0.018 & 0.686 $\pm$ 0.008 \\
    &&\textbf{NMI} &  0.065 $\pm$ 0.002 &  0.076 &       \textbf{0.144 $\pm$ 0.003} &       0.084 $\pm$ 0.011 & --- & 0.097 $\pm$ 0.006 & 0.106 $\pm$ 0.017 & 0.705 $\pm$ 0.008 \\
    &&\textbf{MAP} &  0.188 $\pm$ 0.000 &  0.203 &       \textbf{0.218 $\pm$ 0.000} &       0.202 $\pm$ 0.002 & --- & 0.205 $\pm$ 0.002 & 0.200 $\pm$ 0.004 & 0.831 $\pm$ 0.007 \\
    &&\textbf{MRR} &  0.356 $\pm$ 0.007 &  0.444 &       \textbf{0.462 $\pm$ 0.003} &       0.439 $\pm$ 0.008 & --- & 0.453 $\pm$ 0.003 & 0.391 $\pm$ 0.013 & 0.920 $\pm$ 0.003 \\
    
    \midrule
    \multirow{21}{*}{\textbf{Cars196}} & \multirow{7}{*}{\textbf{Car Model}} & \textbf{MAP@R} &   0.001 $\pm$ 0.000 &  0.235 &     0.374 $\pm$ 0.000 &     0.192 $\pm$ 0.003 & \textbf{0.375 $\pm$ 0.001} & 0.332 $\pm$ 0.002 & 0.200 $\pm$ 0.058 & 0.418 $\pm$ 0.000 \\
    &&\textbf{Prec@1} &   0.011 $\pm$ 0.001 &  0.780 &     \textbf{0.844 $\pm$ 0.001} &     0.729 $\pm$ 0.005 & 0.842 $\pm$ 0.001 & 0.824 $\pm$ 0.002 & 0.638 $\pm$ 0.081 & \textit{0.766 $\pm$ 0.001} \\
    &&\textbf{R-Prec} &   0.010 $\pm$ 0.000 &  0.354 &     0.486 $\pm$ 0.000 &     0.309 $\pm$ 0.005 & \textbf{0.487 $\pm$ 0.001} & 0.450 $\pm$ 0.002 & 0.326 $\pm$ 0.058 & 0.545 $\pm$ 0.000 \\
    
    &&\textbf{AMI} &  -0.000 $\pm$ 0.001 &  0.634 &     0.766 $\pm$ 0.002 &     0.597 $\pm$ 0.010 & \textbf{0.771 $\pm$ 0.002} & 0.738 $\pm$ 0.008 & 0.606 $\pm$ 0.073 & 0.803 $\pm$ 0.000 \\
    &&\textbf{NMI} &   0.142 $\pm$ 0.001 &  0.685 &     0.798 $\pm$ 0.002 &     0.653 $\pm$ 0.008 & \textbf{0.803 $\pm$ 0.002} & 0.774 $\pm$ 0.007 & 0.661 $\pm$ 0.062 & 0.831 $\pm$ 0.000 \\
    &&\textbf{MAP} &   0.011 $\pm$ 0.000 &  0.335 &     0.501 $\pm$ 0.000 &     0.281 $\pm$ 0.005 & \textbf{0.504 $\pm$ 0.000} & 0.456 $\pm$ 0.003 & 0.305 $\pm$ 0.070 & 0.573 $\pm$ 0.000 \\
    &&\textbf{MRR} &   0.047 $\pm$ 0.002 &  0.853 &     \textbf{0.898 $\pm$ 0.000} &     0.815 $\pm$ 0.004 & 0.897 $\pm$ 0.000 & 0.885 $\pm$ 0.001 & 0.745 $\pm$ 0.063 & 0.844 $\pm$ 0.000 \\
    
    \cmidrule{2-11}
    & \multirow{7}{*}{\textbf{Manufacturer}} & \textbf{MAP@R} &  0.005 $\pm$ 0.000 &  0.244 &  \textbf{0.336 $\pm$ 0.001} &  0.212 $\pm$ 0.004 & --- & 0.242 $\pm$ 0.004 & 0.180 $\pm$ 0.022 & 0.514 $\pm$ 0.000 \\
    &&\textbf{Prec@1} &  0.054 $\pm$ 0.003 &  0.890 &  \textbf{0.905 $\pm$ 0.001} &  0.847 $\pm$ 0.008 & --- & 0.855 $\pm$ 0.003 & 0.631 $\pm$ 0.039 & \textit{0.840 $\pm$ 0.001} \\
    &&\textbf{R-Prec} &  0.054 $\pm$ 0.000 &  0.363 &  \textbf{0.445 $\pm$ 0.001} &  0.333 $\pm$ 0.004 & --- & 0.362 $\pm$ 0.004 & 0.309 $\pm$ 0.021 & 0.622 $\pm$ 0.000 \\
    
    &&\textbf{AMI} &  0.001 $\pm$ 0.001 &  0.544 &  \textbf{0.631 $\pm$ 0.002} &  0.509 $\pm$ 0.014 & --- & 0.535 $\pm$ 0.008 & 0.436 $\pm$ 0.026 & 0.725 $\pm$ 0.001 \\
    &&\textbf{NMI} &  0.023 $\pm$ 0.001 &  0.555 &  \textbf{0.640 $\pm$ 0.002} &  0.520 $\pm$ 0.013 & --- & 0.546 $\pm$ 0.008 & 0.449 $\pm$ 0.026 & 0.732 $\pm$ 0.001 \\
    &&\textbf{MAP} &  0.055 $\pm$ 0.000 &  0.358 &  \textbf{0.461 $\pm$ 0.001} &  0.321 $\pm$ 0.005 & --- & 0.355 $\pm$ 0.005 & 0.293 $\pm$ 0.024 & 0.655 $\pm$ 0.000 \\
    &&\textbf{MRR} &  0.155 $\pm$ 0.002 &  0.928 &  \textbf{0.938 $\pm$ 0.001} &  0.899 $\pm$ 0.005 & --- & 0.904 $\pm$ 0.003 & 0.737 $\pm$ 0.030 & 0.891 $\pm$ 0.000 \\
    
    \cmidrule{2-11}
    & \multirow{7}{*}{\textbf{Car Type}} & \textbf{MAP@R} &   0.035 $\pm$ 0.000 &  0.251 &  \textbf{0.361 $\pm$ 0.003} &  0.221 $\pm$ 0.008 & --- & 0.277 $\pm$ 0.006 & 0.244 $\pm$ 0.016 & 0.738 $\pm$ 0.000 \\
    &&\textbf{Prec@1} &   0.173 $\pm$ 0.004 &  \textbf{0.911} &  0.907 $\pm$ 0.002 &  0.883 $\pm$ 0.005 & --- & 0.891 $\pm$ 0.004 & 0.632 $\pm$ 0.031 & \textit{0.891 $\pm$ 0.000} \\
    &&\textbf{R-Prec} &   0.171 $\pm$ 0.000 &  0.407 &  \textbf{0.509 $\pm$ 0.003} &  0.381 $\pm$ 0.008 & --- & 0.437 $\pm$ 0.006 & 0.420 $\pm$ 0.015 & 0.802 $\pm$ 0.000 \\
    
    && \textbf{AMI} &  -0.000 $\pm$ 0.000 &  0.371 &  \textbf{0.479 $\pm$ 0.012} &  0.317 $\pm$ 0.024 & --- & 0.409 $\pm$ 0.011 & 0.390 $\pm$ 0.032 & 0.744 $\pm$ 0.001 \\
    && \textbf{NMI} &   0.001 $\pm$ 0.000 &  0.372 &  \textbf{0.480 $\pm$ 0.012} &  0.318 $\pm$ 0.024 & --- & 0.410 $\pm$ 0.011 & 0.391 $\pm$ 0.032 & 0.744 $\pm$ 0.001 \\
    && \textbf{MAP} &   0.172 $\pm$ 0.000 &  0.413 &  \textbf{0.531 $\pm$ 0.003} &  0.383 $\pm$ 0.008 & --- & 0.446 $\pm$ 0.006 & 0.421 $\pm$ 0.017 & 0.844 $\pm$ 0.000 \\
    && \textbf{MRR} &   0.356 $\pm$ 0.003 &  \textbf{0.946} &  0.942 $\pm$ 0.001 &  0.928 $\pm$ 0.003 & --- & 0.933 $\pm$ 0.002 & 0.753 $\pm$ 0.022 & 0.929 $\pm$ 0.000 \\
    
    \midrule
    \multirow{7}{*}{\textbf{CUB200}} & \multirow{7}{*}{\textbf{Bird Species}} & \textbf{MAP@R} &  0.001 $\pm$ 0.000 &  0.180 &  \textbf{0.265 $\pm$ 0.000} &  0.152 $\pm$ 0.003 & --- & 0.188 $\pm$ 0.002 & 0.151 $\pm$ 0.019 & 0.341 $\pm$ 0.000 \\
    &&\textbf{Prec@1} &  0.012 $\pm$ 0.001 &  0.582 &  \textbf{0.653 $\pm$ 0.001} &  0.526 $\pm$ 0.003 & --- & 0.581 $\pm$ 0.005 & 0.444 $\pm$ 0.036 & \textit{0.653 $\pm$ 0.002} \\
    &&\textbf{R-Prec} &  0.013 $\pm$ 0.000 &  0.297 &  \textbf{0.386 $\pm$ 0.000} &  0.265 $\pm$ 0.004 & --- & 0.306 $\pm$ 0.002 & 0.261 $\pm$ 0.022 & 0.474 $\pm$ 0.000 \\
    
    && \textbf{AMI} &  0.000 $\pm$ 0.002 &  0.562 &  \textbf{0.659 $\pm$ 0.003} &  0.520 $\pm$ 0.009 & --- & 0.578 $\pm$ 0.010 & 0.483 $\pm$ 0.024 & 0.736 $\pm$ 0.002 \\
    && \textbf{NMI} &  0.160 $\pm$ 0.002 &  0.627 &  \textbf{0.711 $\pm$ 0.002} &  0.593 $\pm$ 0.007 & --- & 0.642 $\pm$ 0.008 & 0.564 $\pm$ 0.020 & 0.777 $\pm$ 0.002 \\
    && \textbf{MAP} &  0.015 $\pm$ 0.000 &  0.268 &  \textbf{0.379 $\pm$ 0.000} &  0.235 $\pm$ 0.004 & --- & 0.282 $\pm$ 0.003 & 0.241 $\pm$ 0.023 & 0.488 $\pm$ 0.000 \\
    && \textbf{MRR} &  0.055 $\pm$ 0.001 &  0.704 &  \textbf{0.758 $\pm$ 0.001} &  0.656 $\pm$ 0.002 & --- & 0.702 $\pm$ 0.003 & 0.579 $\pm$ 0.033 & 0.758 $\pm$ 0.001 \\
    
    \midrule
    \multirow{28}{*}{\textbf{DeepFashion}} & \multirow{7}{*}{\textbf{\makecell[r]{Clothing\\Category}}} & \textbf{MAP@R} &   0.023 $\pm$ 0.000 &  0.125 &  \textbf{0.187 $\pm$ 0.001} &  0.113 $\pm$ 0.004 & --- & 0.133 $\pm$ 0.003 & 0.169 $\pm$ 0.018 & 0.322 $\pm$ 0.001 \\
    &&\textbf{Prec@1} &   0.111 $\pm$ 0.004 &  0.452 &  \textbf{0.509 $\pm$ 0.002} &  0.430 $\pm$ 0.006 & --- & 0.455 $\pm$ 0.005 & 0.445 $\pm$ 0.024 & 0.558 $\pm$ 0.006 \\
    &&\textbf{R-Prec} &   0.109 $\pm$ 0.000 &  0.247 &  \textbf{0.322 $\pm$ 0.001} &  0.230 $\pm$ 0.003 & --- & 0.256 $\pm$ 0.003 & 0.302 $\pm$ 0.020 & 0.449 $\pm$ 0.001 \\
    
    && \textbf{AMI} &  -0.001 $\pm$ 0.002 &  0.239 &  \textbf{0.350 $\pm$ 0.003} &  0.228 $\pm$ 0.010 & --- & 0.266 $\pm$ 0.009 & 0.297 $\pm$ 0.027 & 0.439 $\pm$ 0.001 \\
    && \textbf{NMI} &   0.049 $\pm$ 0.002 &  0.276 &  \textbf{0.383 $\pm$ 0.003} &  0.266 $\pm$ 0.009 & --- & 0.303 $\pm$ 0.009 & 0.333 $\pm$ 0.026 & 0.467 $\pm$ 0.001 \\
    && \textbf{MAP} &   0.111 $\pm$ 0.000 &  0.226 &  \textbf{0.307 $\pm$ 0.001} &  0.213 $\pm$ 0.003 & --- & 0.238 $\pm$ 0.004 & 0.290 $\pm$ 0.020 & 0.449 $\pm$ 0.001 \\
    && \textbf{MRR} &   0.242 $\pm$ 0.004 &  0.588 &  \textbf{0.631 $\pm$ 0.002} &  0.565 $\pm$ 0.004 & --- & 0.585 $\pm$ 0.004 & 0.577 $\pm$ 0.022 & 0.668 $\pm$ 0.003 \\
    
    \cmidrule{2-11}
    & \multirow{7}{*}{\textbf{Texture}} & \textbf{MAP@R} &  0.118 $\pm$ 0.000 &  0.187 &  \textbf{0.330 $\pm$ 0.004} &  0.112 $\pm$ 0.004 & --- & 0.222 $\pm$ 0.005 & 0.163 $\pm$ 0.007 & 0.661 $\pm$ 0.001 \\
    &&\textbf{Prec@1} &  0.296 $\pm$ 0.007 &  0.602 &  \textbf{0.668 $\pm$ 0.003} &  0.433 $\pm$ 0.005 & --- & 0.612 $\pm$ 0.007 & 0.438 $\pm$ 0.017 & 0.806 $\pm$ 0.003 \\
    &&\textbf{R-Prec} &  0.294 $\pm$ 0.000 &  0.358 &  \textbf{0.480 $\pm$ 0.004} &  0.229 $\pm$ 0.004 & --- & 0.388 $\pm$ 0.004 & 0.296 $\pm$ 0.008 & 0.743 $\pm$ 0.000 \\
    
    && \textbf{AMI} &  0.000 $\pm$ 0.000 &  0.081 &  \textbf{0.305 $\pm$ 0.014} & 0.224 $\pm$ 0.005 & --- & 0.143 $\pm$ 0.012 & 0.295 $\pm$ 0.014 & 0.551 $\pm$ 0.001 \\
    && \textbf{NMI} &  0.003 $\pm$ 0.000 &  0.083 &  \textbf{0.307 $\pm$ 0.014} & 0.262 $\pm$ 0.004 & --- & 0.145 $\pm$ 0.012 & 0.330 $\pm$ 0.013 & 0.553 $\pm$ 0.001 \\
    && \textbf{MAP} &  0.295 $\pm$ 0.000 &  0.363 &  \textbf{0.496 $\pm$ 0.004} & 0.211 $\pm$ 0.004 & --- & 0.395 $\pm$ 0.005 & 0.282 $\pm$ 0.009 & 0.767 $\pm$ 0.000 \\
    && \textbf{MRR} &  0.479 $\pm$ 0.006 &  0.723 &  \textbf{0.768 $\pm$ 0.001} & 0.568 $\pm$ 0.003 & --- & 0.728 $\pm$ 0.004 & 0.573 $\pm$ 0.016 & 0.865 $\pm$ 0.002 \\
    
    \cmidrule{2-11}
    & \multirow{7}{*}{\textbf{Fabric}} & \textbf{MAP@R} &  0.324 $\pm$ 0.000 &  0.340 &  \textbf{0.377 $\pm$ 0.002} &  0.108 $\pm$ 0.003 & --- & 0.356 $\pm$ 0.003 & 0.172 $\pm$ 0.006 & 0.642 $\pm$ 0.003 \\
    &&\textbf{Prec@1} &  0.494 $\pm$ 0.006 &  0.645 &  \textbf{0.661 $\pm$ 0.006} &  0.426 $\pm$ 0.007 & --- & 0.650 $\pm$ 0.006 & 0.447 $\pm$ 0.019 & 0.778 $\pm$ 0.004 \\
    &&\textbf{R-Prec} &  0.498 $\pm$ 0.000 &  0.526 &  \textbf{0.560 $\pm$ 0.002} &  0.224 $\pm$ 0.004 & --- & 0.539 $\pm$ 0.002 & 0.307 $\pm$ 0.005 & 0.735 $\pm$ 0.002 \\
    
    && \textbf{AMI} &  0.000 $\pm$ 0.000 &  0.049 &  \textbf{0.119 $\pm$ 0.004} & 0.219 $\pm$ 0.012 & --- & 0.079 $\pm$ 0.012 & 0.302 $\pm$ 0.010 & 0.403 $\pm$ 0.006 \\
    && \textbf{NMI} &  0.003 $\pm$ 0.000 &  0.051 &  \textbf{0.121 $\pm$ 0.004} & 0.257 $\pm$ 0.011 & --- & 0.081 $\pm$ 0.012 & 0.337 $\pm$ 0.010 & 0.405 $\pm$ 0.006 \\
    && \textbf{MAP} &  0.499 $\pm$ 0.000 &  0.524 &  \textbf{0.556 $\pm$ 0.002} & 0.208 $\pm$ 0.004 & --- & 0.536 $\pm$ 0.003 & 0.294 $\pm$ 0.004 & 0.746 $\pm$ 0.002 \\
    && \textbf{MRR} &  0.636 $\pm$ 0.004 &  0.764 &  \textbf{0.775 $\pm$ 0.003} & 0.564 $\pm$ 0.005 & --- & 0.767 $\pm$ 0.004 & 0.580 $\pm$ 0.015 & 0.848 $\pm$ 0.003 \\
    
    \cmidrule{2-11}
    & \multirow{7}{*}{\textbf{Fit}} & \textbf{MAP@R} &   0.518 $\pm$ 0.000 &  0.533 &  \textbf{0.539 $\pm$ 0.004} &  0.111 $\pm$ 0.010 & --- & 0.534 $\pm$ 0.003 & 0.161 $\pm$ 0.018 & 0.820 $\pm$ 0.001 \\
    &&\textbf{Prec@1} &   0.666 $\pm$ 0.006 &  \textbf{0.771} &  0.765 $\pm$ 0.004 &  0.431 $\pm$ 0.005 & --- & 0.767 $\pm$ 0.007 & 0.429 $\pm$ 0.019 & 0.878 $\pm$ 0.006 \\
    &&\textbf{R-Prec} &   0.666 $\pm$ 0.000 &  0.675 &  \textbf{0.680 $\pm$ 0.001} &  0.227 $\pm$ 0.009 & --- & 0.677 $\pm$ 0.001 & 0.294 $\pm$ 0.017 & 0.871 $\pm$ 0.001 \\
    
    && \textbf{AMI} &  -0.000 $\pm$ 0.000 &  0.002 &  0.003 $\pm$ 0.001 &  0.217 $\pm$ 0.008 & --- & \textbf{0.013 $\pm$ 0.004} & 0.284 $\pm$ 0.017 & 0.376 $\pm$ 0.002 \\
    && \textbf{NMI} &   0.000 $\pm$ 0.000 &  0.002 &  \textbf{0.004 $\pm$ 0.001} &  0.255 $\pm$ 0.008 & --- & \textbf{0.013 $\pm$ 0.004} & 0.320 $\pm$ 0.016 & 0.377 $\pm$ 0.002 \\
    && \textbf{MAP} &   0.667 $\pm$ 0.000 &  0.685 &  \textbf{0.689 $\pm$ 0.002} &  0.211 $\pm$ 0.009 & --- & 0.685 $\pm$ 0.002 & 0.320 $\pm$ 0.016 & 0.879 $\pm$ 0.001 \\
    && \textbf{MRR} &   0.772 $\pm$ 0.005 &  \textbf{0.850} &  0.846 $\pm$ 0.003 &  0.566 $\pm$ 0.003 & --- & 0.845 $\pm$ 0.003 & 0.565 $\pm$ 0.017 & 0.919 $\pm$ 0.003 \\
    
    \midrule
    \multirow{14}{*}{\textbf{Movie Posters}} & \multirow{7}{*}{\textbf{Genre}} & \textbf{MAP@R} &  0.041 $\pm$ 0.000 &  0.114 &  \textbf{0.149 $\pm$ 0.000} &  0.091 $\pm$ 0.003 & --- & 0.084 $\pm$ 0.001 & 0.098 $\pm$ 0.024 & 0.196 $\pm$ 0.001 \\
    &&\textbf{Prec@1} &  0.175 $\pm$ 0.004 &  0.418 &  \textbf{0.440 $\pm$ 0.002} &  0.381 $\pm$ 0.007 & --- & 0.366 $\pm$ 0.004 & 0.333 $\pm$ 0.030 & \textit{0.432 $\pm$ 0.007} \\
    &&\textbf{R-Prec} &  0.174 $\pm$ 0.000 &  0.273 &  \textbf{0.306 $\pm$ 0.000} &  0.246 $\pm$ 0.003 & --- & 0.237 $\pm$ 0.001 & 0.248 $\pm$ 0.031 & 0.364 $\pm$ 0.001 \\
    
    && \textbf{AMI} &  0.000 $\pm$ 0.000 &  0.186 &  \textbf{0.196 $\pm$ 0.003} &  0.150 $\pm$ 0.004 & --- & 0.101 $\pm$ 0.007 & 0.107 $\pm$ 0.044 & 0.254 $\pm$ 0.001 \\
    && \textbf{NMI} &  0.013 $\pm$ 0.000 &  0.196 &  \textbf{0.206 $\pm$ 0.003} &  0.160 $\pm$ 0.004 & --- & 0.112 $\pm$ 0.007 & 0.118 $\pm$ 0.043 & 0.263 $\pm$ 0.001 \\
    && \textbf{MAP} &  0.175 $\pm$ 0.000 &  0.261 &  \textbf{0.298 $\pm$ 0.000} &  0.236 $\pm$ 0.003 & --- & 0.227 $\pm$ 0.001 & 0.242 $\pm$ 0.028 & 0.354 $\pm$ 0.001 \\
    && \textbf{MRR} &  0.346 $\pm$ 0.005 &  0.573 &  \textbf{0.587 $\pm$ 0.001} &  0.540 $\pm$ 0.005 & --- & 0.529 $\pm$ 0.004 & 0.495 $\pm$ 0.027 & 0.579 $\pm$ 0.006 \\
    
    \cmidrule{2-11}
    & \multirow{7}{*}{\textbf{\makecell[r]{Production\\Country}}} & \textbf{MAP@R} &   0.446 $\pm$ 0.000 &  0.493 &  \textbf{0.513 $\pm$ 0.001} &  0.489 $\pm$ 0.004 & --- & 0.477 $\pm$ 0.002 & 0.494 $\pm$ 0.007 & 0.581 $\pm$ 0.000 \\
    &&\textbf{Prec@1} &   0.592 $\pm$ 0.005 &  0.693 &  \textbf{0.698 $\pm$ 0.003} &  0.679 $\pm$ 0.007 & --- & 0.681 $\pm$ 0.003 & 0.649 $\pm$ 0.007 & 0.718 $\pm$ 0.003 \\
    &&\textbf{R-Prec} &   0.592 $\pm$ 0.000 &  0.625 &  \textbf{0.639 $\pm$ 0.001} &  0.621 $\pm$ 0.002 & --- & 0.613 $\pm$ 0.001 & 0.624 $\pm$ 0.006 & 0.693 $\pm$ 0.000 \\
    
    && \textbf{AMI} &  -0.000 $\pm$ 0.001 &  0.063 &  \textbf{0.076 $\pm$ 0.001} &  0.057 $\pm$ 0.002 & --- & 0.048 $\pm$ 0.002 & 0.047 $\pm$ 0.007 & 0.110 $\pm$ 0.001 \\
    && \textbf{NMI} &   0.046 $\pm$ 0.001 &  0.106 &  \textbf{0.118 $\pm$ 0.001} &  0.100 $\pm$ 0.002 & --- & 0.093 $\pm$ 0.002 & 0.091 $\pm$ 0.007 & 0.150 $\pm$ 0.001 \\
    && \textbf{MAP} &   0.592 $\pm$ 0.000 &  0.634 &  \textbf{0.648 $\pm$ 0.001} &  0.629 $\pm$ 0.003 & --- & 0.620 $\pm$ 0.001 & 0.629 $\pm$ 0.005 & 0.692 $\pm$ 0.000 \\
    && \textbf{MRR} &   0.692 $\pm$ 0.003 &  0.770 &  \textbf{0.773 $\pm$ 0.001} &  0.760 $\pm$ 0.003 & --- & 0.760 $\pm$ 0.001 & 0.739 $\pm$ 0.005 & 0.788 $\pm$ 0.002 \\
    
    \bottomrule
    \end{tabular}
}
\end{table}

In \Cref{tab:results}, we show all evaluation metric results for our experiments.
Namely, these are Mean Average Precision at R (MAP@R)~\cite{musgrave2020metric}, Precision at 1 (Prec@1), R-Precision (R-Prec), Adjusted Mutual Information (AMI), Normalized Mutual Information (NMI), Mean Average Precision (MAP), and Mean Reciprocal Rank (MRR).
For a overview of different evaluation metrics in the context of Deep Metric Learning, we refer the reader to the appendix of the paper by Roth et al.~\cite{roth2020revisiting}.

\section{Details of Datasets and Similarity Notions}

We experiment with five datasets and overall thirteen similarity notions.
In the following, we give more insights into the datasets, their similarity notions, and how we obtained aspects that were embedded in the string templates.

\subsection{Synthetic Cars~\cite{kobs2021different}}

This dataset contains 3D-rendered car images with different car models, car colors, background colors, car orientations, sun positions, and camera angles, all sampled independently at random.
We use the first \num{1000} images to speed up evaluation (not training, since no images are used for training).

Since the images have annotated image properties, we are able to use different properties as different similarity notions.
Note that we do not use the provided labels for training, only for evaluation.
We use the following similarity notions and corresponding text prompts:

\begin{itemize}
    \item \textbf{Car model} (``Two car images are similar if they show the same car model''): The dataset provides six different car models. To train the dimensionality reduction, we use a list of car models scraped from an online car dealer.\footnote{\url{https://www.kbb.com/car-make-model-list/new/view-all/make/}} Each of the \num{569} car model's name is then embedded in the text prompt template ``a photo of a [car model]''.
    \item \textbf{Car color} (``Two car images are similar if both cars have the same color''): The dataset samples the car and background colors uniformly from the hue, saturation, and value (HSV) color space. To get binary similarities for evaluation, we find the nearest CSS2.1 color name (overall \num{18} possible colors, e.g. ``orange'', ``black'') for each HSV color. We use all color names in the string template ``a [color name] car'' as text prompts.
    \item \textbf{Background color} (``Two images are similar if they show the same background color''): We use the same process as for the car color but only change the text prompt template to ``a car in front of a [color] background''.
\end{itemize}

\subsection{Cars196~\cite{KrauseStarkDengFei-Fei_3DRR2013}}

Cars196 is a common dataset in Deep Metric Learning, which features \num{16185} real world car images.
Usually, it is split into \num{196} classes, each one representing images of one car model.
As commonly done in Deep Metric Learning papers, we use the second half of the classes (\num{8131} images) for the evaluation to be able to compare our method to methods from the literature that are trained explicitly on the training split of the dataset.

The following similarity notions and string templates are used:
\begin{itemize}
    \item \textbf{Car model} (``Two car images are similar if they show the same car model''): The default definition for this dataset. We use the same list of car models and the same text prompt template as for the synthetic car dataset.
    \item \textbf{Manufacturer} (``Two car images are similar if both cars have the same manufacturer''): This is a superset of classes from the car model definition, i.e. the multiple car models belong to one manufacturer. In the test dataset, there are \num{35} different car manufacturers. We use the template ``a photo of a car produced by [manufacturer]'' with all \num{46} manufacturers extracted from the same website as for the car models.
    \item \textbf{Car type} (``Two car images are similar if both cars have the same car type''): Car types like convertibles, SUV's etc. are coming from different manufacturers, but usually look similar. Cars196's dataset has seven different car types, which are also used for prompting, since there are only a certain amount of car types. They are embedded in the template string ``a photo of a [car type]''.
\end{itemize}
    
\subsection{CUB200~\cite{WahCUB_200_2011}}

CUB200 is a commonly used dataset in Deep Metric Learning, consisting of images showing birds, usually grouped by bird species.
While the dataset has \num{200} classes, we again use the second half of classes for evaluation.
Due to the lack of additional metadata for each image, we only use the default similarity notion for evaluation:
\begin{itemize}
    \item \textbf{Bird species} (``Two bird images are similar if they show the same bird species''): As text prompts, we use the very generic ``a photo of a [bird species]'' with all bird species used in the training dataset. This ensures that the test class names are not used in our method.
\end{itemize}

\subsection{DeepFashion~\cite{liu2016deepfashion}}

The dataset contains images of persons wearing different clothes.
It has \num{4000} test images we use for evaluation.
The similarity notions and the corresponding text prompts for training are:
\begin{itemize}
    \item \textbf{Category} (``Two clothing images are similar if they show the same type of clothing''): \num{50} categories are available in the dataset (e.g. ``Anorak'', ``Turtleneck''). We use all categories in our text prompts with template ``a photo of a person wearing a [clothing category]''.
    \item \textbf{Texture} (``Two clothing images are similar if they share the same texture''): There are seven different texture types in the dataset (e.g. ``striped''). We use all of them for our prompts with template ``a photo of a person wearing clothes with a [texture type] texture''.
    \item \textbf{Fabrics} (``Two clothing images are similar if they use the same kind of fabric''): We use all six different fabric types (e.g. ``cotton'') in the template ``a photo of a person wearing clothes made out of [fabric type]''.
    \item \textbf{Fit} (``Two clothing images are similar if they have the same fit''): We use all three fit types (``tight'', ``loose'', ``conventional'') in ``a photo of a person wearing clothes with a [fit type] fit''.
\end{itemize}

\subsection{Movie Posters~\cite{chu2017movie}}

This is a dataset of movie posters and corresponding metadata about the movie.
We overall are able to read \num{8052} different movie posters and use them in our experiments.
While this dataset is not a commonly used dataset in Deep Metric Learning, it still can be used in our setting and with an interesting task: Finding similar movie posters and thus movies based on the desired similarity notion.

We use the following definitions and prompt templates:
\begin{itemize}
    \item \textbf{Genre} (``Two movie posters are similar if both films share the same genre''): This similarity notion assumes that there are visual clues in the movie posters that indicate the genre. We argue that this is the case, at least for certain genres, such as action movies, where the protagonist is often shown with a gun while looking serious. There are \num{25} genres (each movie can have multiple genres, so we only take the main one) in the dataset that we use in the string template ``a poster of a [genre] movie''.
    \item \textbf{Production country} (``Two movie posters are similar if both films were mainly produced in one country''): There are \num{69} different production countries listed for the dataset (we again use only the main one if there are multiple for one movie). We use all of these countries in the string template ``a poster of a movie produced in [country]''. Again, the task assumes that the main production country is somehow visible in the movie poster, which is usually true for, for example, movies from the USA and India.
\end{itemize}

\section{Saliency Maps}

\begin{figure}[t]
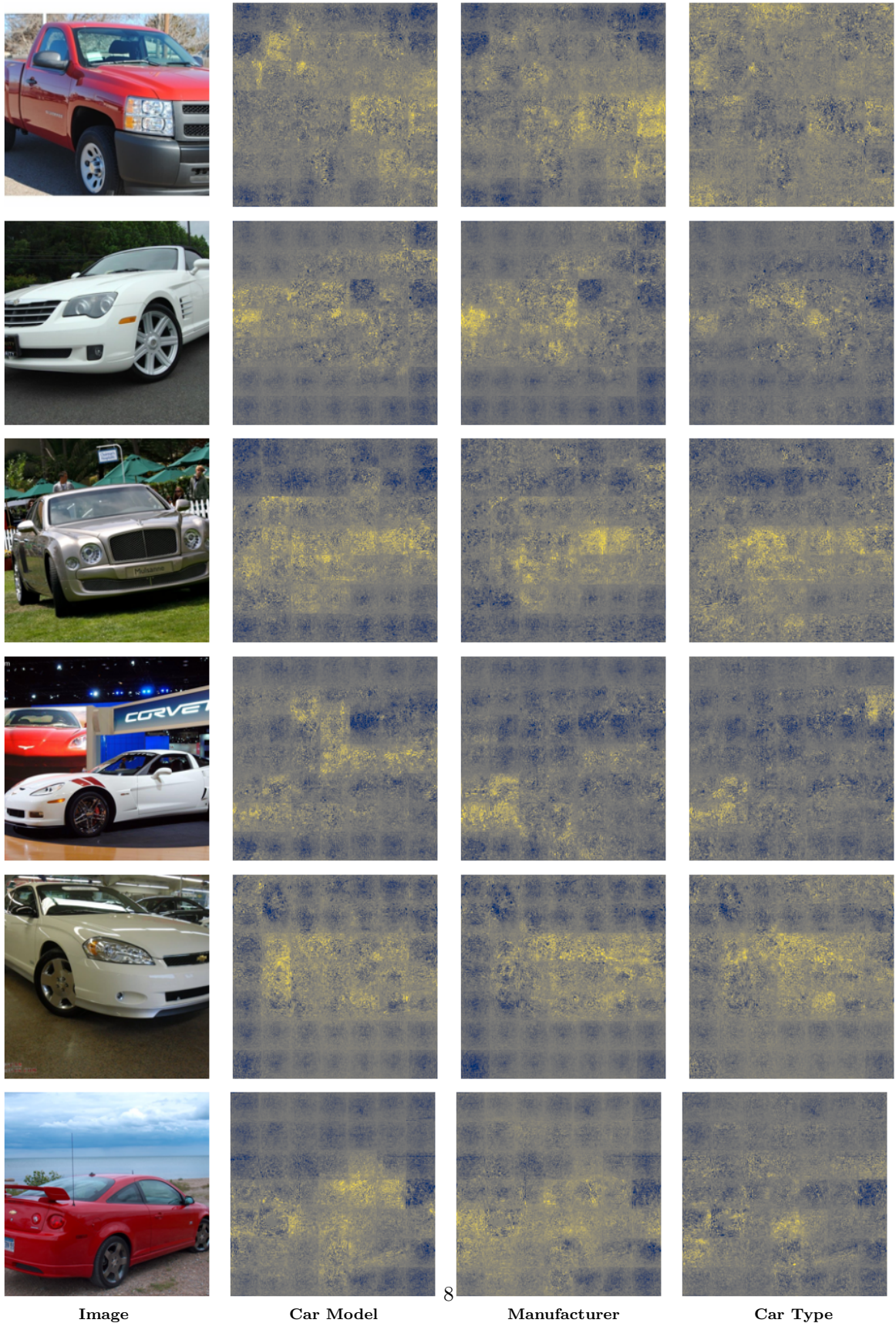

    \centering
    
    \foreach \id in {006091,006519,003320,004578,005803,005296}{
        \includegraphics[width=0.24\textwidth]{images/explainer/\id_image.png}
        \hfill
        \includegraphics[width=0.24\textwidth]{images/explainer/\id_Car_Model.png}
        \hfill
        \includegraphics[width=0.24\textwidth]{images/explainer/\id_Manufacturer.png}
        \hfill
        \includegraphics[width=0.24\textwidth]{images/explainer/\id_Car_Type.png}
    }
    
    \vspace{-6mm}
    \begin{multicols}{4}
        \textbf{\scriptsize Image}\\
        \textbf{\scriptsize Car Model}\\
        \textbf{\scriptsize Manufacturer}\\
        \textbf{\scriptsize Car Type}
    \end{multicols}
    \vspace{-7mm}
    
    \caption{Randomly chosen example images from the Cars196 dataset and the differences in saliency maps between each similarity notion and CLIP.
    Yellow regions denote that \model pays more attention to that region than CLIP.}
    \label{fig:explainer_analysis}
\end{figure}

\Cref{fig:explainer_analysis} shows six randomly chosen example images from the Cars196~\cite{KrauseStarkDengFei-Fei_3DRR2013} dataset.
As in the main paper, we compute the saliency maps for CLIP and each of \model' similarity notions and visualize their difference.

\section{Embedding Space Visualizations}

We visualize the embeddings produces by \model for multiple similarity notions of the Cars196 dataset using TriMap~\cite{2019TRIMAP}.
In addition to the ``Car Model'', ``Car Manufacturer'', and ``Car Type'' similarity notions, we also add the ``Car Color'' similarity notion, which is not present in the original dataset's metadata.
For visualization purposes only, we rudimentary label each image with one of eight colors (`black', `blue', `white', `yellow', `silver', `red', `mixed', `other').
Note that since \model does not need labeled images, this process was only necessary for this visualization of the embedding space.
The visualizations still show that cars with the same properties are clustered relatively well, even though \model does not use any training images but only text prompts.

\begin{figure}[t]
    \centering
    
    \includegraphics[width=0.45\textwidth]{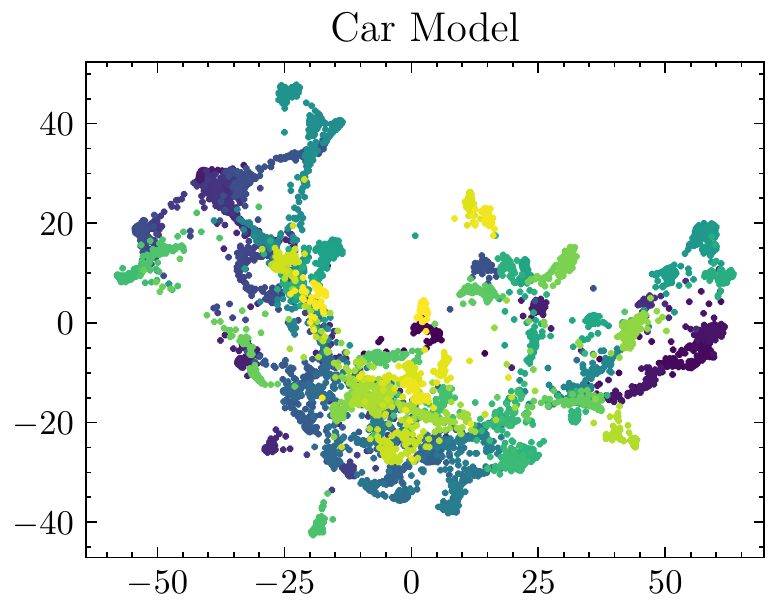}
    \hfill
    \includegraphics[width=0.45\textwidth]{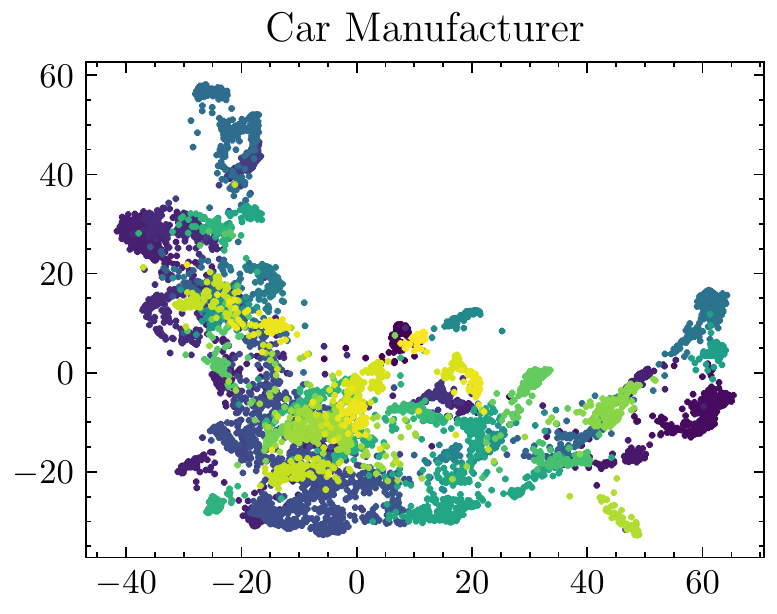}
    
    \includegraphics[width=0.45\textwidth]{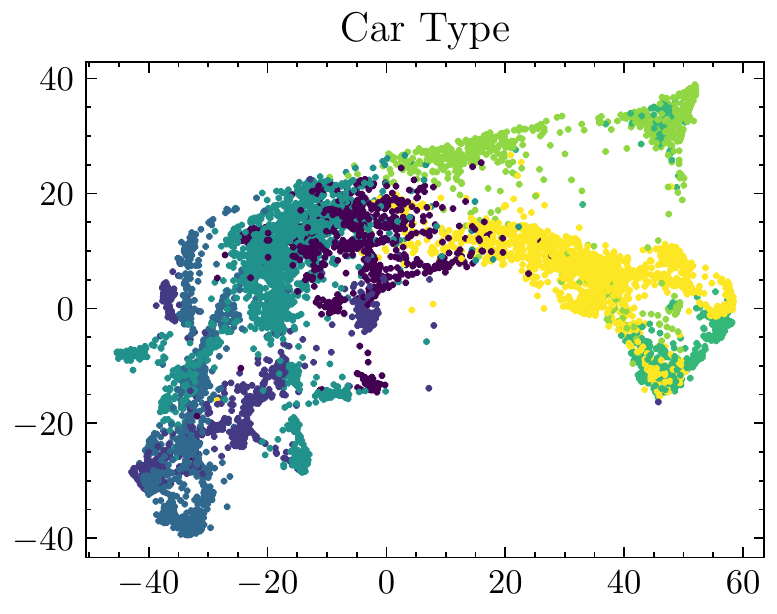}
    \hfill
    \includegraphics[width=0.45\textwidth]{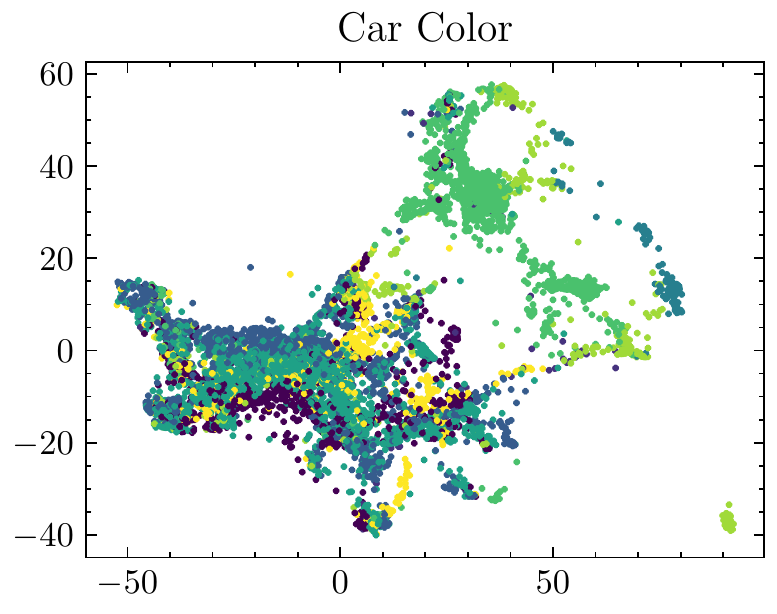}

    \caption{TriMap visualizations for multiple similarity notions of the Cars196 dataset.}
    \label{fig:explainer_analysis}
\end{figure}

\bibliographystyle{plain}
\bibliography{supplement}